\documentclass{ieeetj}
\usepackage{cite}
\usepackage{amsmath,amssymb,amsfonts}
\usepackage{algorithmic}
\usepackage{graphicx,color}
\usepackage{textcomp}
\usepackage[caption=false,font=footnotesize]{subfig}
\usepackage{xcolor}
\usepackage{hyperref}
\hypersetup{hidelinks=true}
\usepackage{algorithm,algorithmic}
\usepackage[capitalise]{cleveref}
\usepackage{multirow}
\usepackage{booktabs}

\usepackage{colortbl}

\definecolor{lightgray}{gray}{0.9}

\def\BibTeX{{\rm B\kern-.05em{\sc i\kern-.025em b}\kern-.08em
    T\kern-.1667em\lower.7ex\hbox{E}\kern-.125emX}}
\AtBeginDocument{\definecolor{tmlcncolor}{cmyk}{0.59,0.94,0.01,0.02}\definecolor{NavyBlue}{RGB}{0,86,125}}

\def\OJlogo{\vspace{-4pt}~}
\def\seclogo{\vspace{10pt}~}

\def\authorrefmark#1{\ensuremath{^{\textbf{#1}}}}

\begin{document}
\receiveddate{June 26, 2025}
\reviseddate{January 29, 2026}
\accepteddate{February 5, 2026}
\publisheddate{---}
\currentdate{\today}


\title{Digging for Data: Experiments in Rock Pile Characterization Using Only Proprioceptive Sensing in Excavation}

\author{Unal Artan\authorrefmark{1}, Martin Magnusson\authorrefmark{1}, and Joshua A.\ Marshall\authorrefmark{2}, Senior Member, IEEE}
\affil{Center for Applied Autonomous Sensor Systems, \"{O}rebro University, SE-701 82 \"{O}rebro, Sweden}
\affil{Ingenuity Labs Research Institute, Smith Engineering, Queen's University, Kingston, ON K7L 3N6 Canada}
\corresp{Corresponding author: Unal Artan (email: unal.artan@oru.se).}
\authornote{This work was carried out within the Impact Innovation program Swedish Metals \& Minerals under grant 2024-02697, a joint effort of the Swedish Energy Agency, Formas, and Vinnova, and by Epiroc Rock Drills AB. Additional support was provided by the Knowledge Foundation (KK-stiftelsen) through the Synergy project TeamRob under grant 20210016. Early development of the concepts presented in this paper was supported by the NSERC Canadian Robotics Network (NCRN) under grant NETGP 508451-17.}

\begin{abstract}
Characterization of fragmented rock piles is a fundamental task in the mining and quarrying industries, where rock is fragmented by blasting, transported using wheel loaders, and then sent for further processing.  This field report studies a novel method for estimating the relative particle size of fragmented rock piles from only proprioceptive data collected while digging with a wheel loader. Rather than employ exteroceptive sensors (e.g., cameras or LiDAR sensors) to estimate rock particle sizes, the studied method infers rock fragmentation from an excavator's inertial response during excavation. This paper expands on research that postulated the use of wavelet analysis to construct a unique feature that is proportional to the level of rock fragmentation. We demonstrate through extensive field experiments that the ratio of wavelet features, constructed from data obtained by excavating in different rock piles with different size distributions, approximates the ratio of the mean particle size of the two rock piles. Full-scale excavation experiments were performed with a battery electric, 18-tonne capacity, load-haul-dump (LHD) machine in representative conditions in an operating quarry. The relative particle size estimates generated with the proposed sensing methodology are compared with those obtained from both a vision-based fragmentation analysis tool and from sieving of sampled materials.
\end{abstract}



\maketitle

\section{INTRODUCTION}

\begin{figure}
    \centering
    \includegraphics[width=\linewidth]{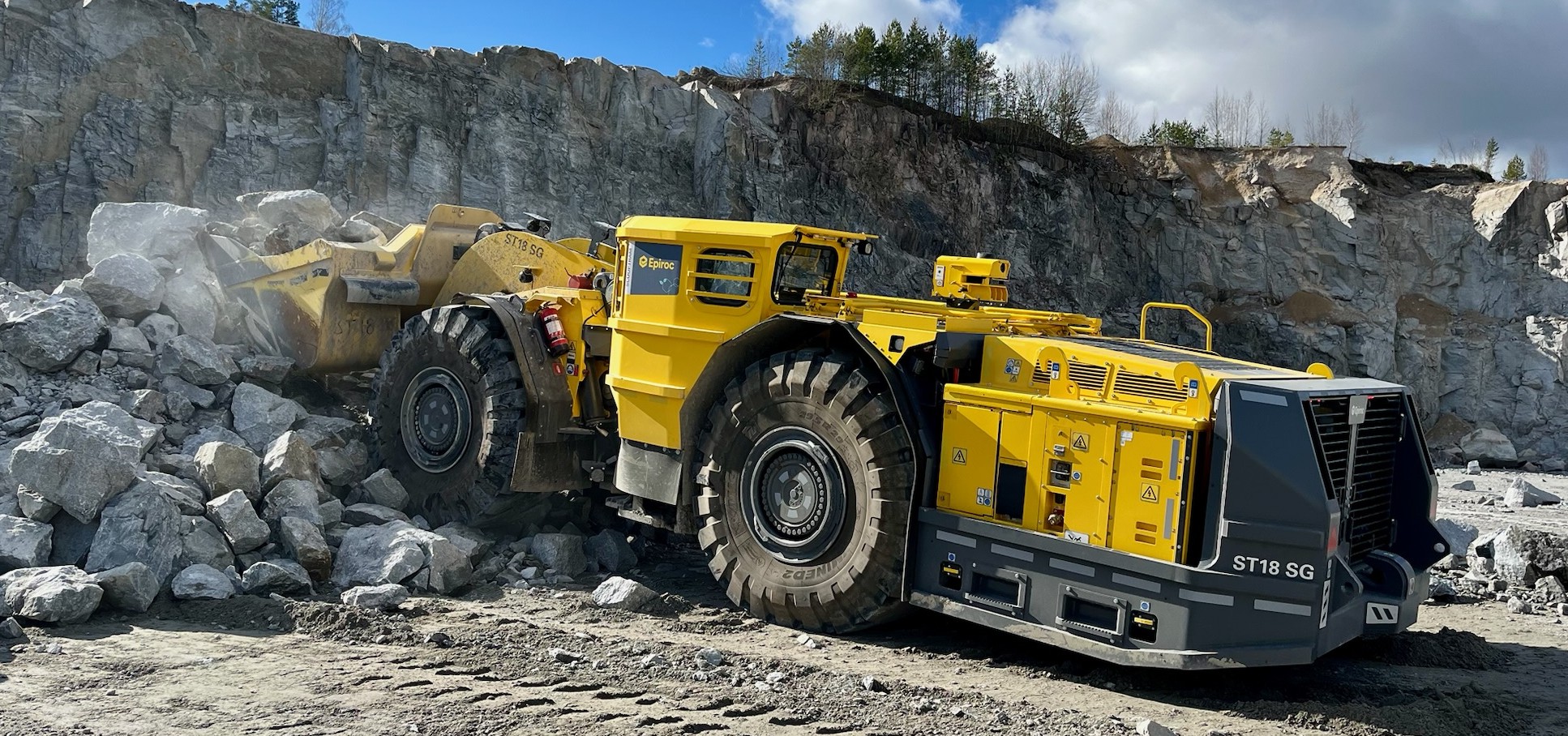}
    \caption{Example excavation operation where the battery electric load-haul-dump (LHD) used in the study presented by this field report excavates fragmented rock into its bucket at an operating quarry.}
    \label{fig:material-handling-example}
\end{figure}

\IEEEPARstart{M}{aterial} handling activities like haulage and excavation, as depicted in Figure\ \ref{fig:material-handling-example}, are prevalent in the aggregate, construction, and mining industries. The materials are often broken rocks created by blasting, hammering, or crushing. Bulk properties such as density, water content, texture, and size distribution are commonly used to distinguish different materials. Size distribution, in particular, is used to distinguish aggregates used in construction\ \cite{astm-mnl32,OPSS1010,ISO-6274:1982,SS-EN-933} and is also a key factor for ensuring efficiency in downstream processes in the mining industry. Given the increasing scarcity of raw materials and the high costs of material handling tasks, quality data---such as rock fragmentation information---is of critical significance for optimizing and automating these activities.  

This field report studies a new approach for rock fragmentation analysis that automatically estimates relative fragmentation by using proprioceptive data collected by an onboard inertial measurement unit (IMU) while digging with a wheel loader.  This is in contrast to mainstream approaches that use vision-based systems and image processing or expensive and limited laboratory testing.  Based on the extensive, full-scale fieldwork presented in this paper, we also contend that the studied method is particularly well suited to operations that employ robotic excavation technology.

Obtaining size distribution estimates can be a time consuming and difficult task.  In the mining industry, drilling and blasting operations target a desired size distribution \cite{KuznetsovV.M.1973Tmdo,Cunningham2005,SanchidrianJoseA.2017Addo,BEYGLOU2017}, which is also an input to crushing and grinding circuits. Significant upfront capital is required for the design and implementation of crushing and grinding circuits\ \cite{alma994119063405158}. Deviation from these design parameters can lead to suboptimal operations~\cite{ZhouPing2009IOCf,BENGTSSON2017,DunneRobertC2019MO}. Because crushing and grinding, for example, consume a significant amount of an operation's energy\ \cite{Jeswiet2016,MarcAllen-energyConsumption}, suboptimal performance is highly undesirable.

Moreover, especially in underground mines, it is difficult to capture a representative sample of the run-of-mine material safely and without disrupting operations. Current methods for estimating size distribution vary from qualitative assessments, sorting using different sized meshes\ \cite{astm-mnl32,ISO-6274:1982}, and obtaining image(s) that are processed to identify the boundary of different rocks in the image\ \cite{WipFrag,Bamford2017,ThurleyMatthewJ2013AISa}. Each of these approaches have their own benefits and drawbacks.  

This paper studies a new approach to fragmentation estimation that leverages the interactions between excavation equipment and fragmented rock by using only proprioceptive sensing.  This approach has its roots in concepts utilized in material testing to extract material hardness of rocks using, for example, the Split Hopkinson bar \cite{alma9927930373405158}.  Signal amplitudes due to strikes are assessed and correlated to the sample material's properties.  In material handling, the signals capturing machine-material interactions contain information about the rock pile's characteristics (e.g., mean particle size).  Inference can be made on rock pile properties from changes in the acquired signal. For example, deciding whether two rock piles are similar or different in size by evaluating the two signals obtained by physically interacting with the rock piles and measuring the excavator's inertial response.  In this paper, we study the feasibility and practicality of using wavelet analysis to construct a unique feature that is proportional to the level of rock fragmentation.

\subsection{TOWARDS AUTONOMOUS EXCAVATION}

A challenge in autonomous excavation is that of adapting to variations in the excavation media. The ability to detect material characteristics enables autonomous systems to adjust their behaviour in real-time, which is essential for robust automated loading across varying conditions. Recent research has shown that adaptive controllers for autonomous excavation require continuous feedback about material properties to optimize their behaviour and avoid failure modes such as stalling or incomplete bucket filling \cite{egli2024reinforcement, jud2017planning, sandzimier2020data}.

Reinforcement learning-based excavation controllers adapt online to varying soil conditions by observing proprioceptive signals such as joint torques and hydraulic pressures as implicit indicators of material properties \cite{egli2024reinforcement, egli2022soil, dadhich2020adaptation, huang2021data, halbach2019neural, yang2020learning, yang2021neural, strokina2022visual}. Model predictive control and trajectory optimization approaches similarly require accurate material property estimation to generate feasible trajectories \cite{jud2017planning, sandzimier2020data, makipenttila2024autonomous}. The proprioceptive signals used for material classification in this paper---accelerations from bucket-rock collisions, and hydraulic pressures---are the same signals these autonomous systems rely on for adaptive control.

Demonstrations on full-scale equipment in realistic operating conditions is critical for practical deployment. Most autonomous excavation research is conducted on smaller research platforms or in simulation environments \cite{terenzi2025autonomous}. The proprioceptive signals analyzed in this work are the same signals used by these controllers, and the classification demonstrated in this paper represents a necessary first step towards integration with closed-loop control. However, further investigation remains to verify the applicability to autonomous excavation of the specific wavelet features studied by the work presented in this paper.

\subsection{ABOUT THIS PAPER}

By way of full-scale experiments at an operating quarry, this field report expands on preliminary research suggesting that proprioceptive measurements capturing equipment interacting with excavation media (e.g., a pile of fragmented rocks) can be processed using wavelet analysis to automatically uncover information about the level of material fragmentation, specifically the mean particle size\ \cite{ArtanU-PhD2022}.  

The paper first describes different approaches to characterizing fragmented rock and provides an overview of wavelet analysis and unique features. The experimental setup is described next.  We evaluate material characterization by assessing relative mean particle sizes using standard fragmentation analysis tools and those from the developed method by processing signals from different proprioceptive sensors to characterize five different rock piles, followed by a discussion of the results.

\section{ROCK PILE CHARACTERIZATION}
\label{sec:fragmented-rock}

This section provides some background and terminology about approaches to modeling rock pile characteristics that are commonly used in industry.  We also provide a brief introduction to other techniques used for assessing rock pile fragmentation, including vision-based approaches that use image processing, as well as sieve analysis that depends on physical sampling and direct measurements.

\subsection{SIZE DISTRIBUTION MODELS}
\label{sec:rock_pile_models}

Suppose a fragmented rock pile consists of $N$ ``cubes'' with size $d_{i} = (0,d_{\max}]$ where, $d_{\max}$ is the largest rock size in the pile.  Assume that all of the rocks have the same density $\rho$, and the mass of a rock $i$, with size $d_{i}$, is $m_{i} = \rho d^{3}_{i}$.  The total mass of the rock pile composed of $N$ rocks is then 
\begin{equation}
    M = \sum_{i=1}^{N} m_{i} = \rho \sum_{i=1}^{N} d^{3}_{i}.
\end{equation}
The mean particle size, $\bar{d}$, of the rock pile is therefore
\begin{equation}
    \bar{d} = \dfrac{1}{N}\sum_{i=1}^{N} d_{i}.
    \label{eq:average_particle_size}
\end{equation}

However, in practice, measuring individual particles is impractical.  Instead, rock piles are usually are described by using their cumulative mass fraction, $P(x)$, which is the mass fraction of particles of size less than or equal to a particular size $x$.  Define
\begin{equation}
    M_{\leq x} = \sum_{i:d_i\leq x} m_i = \rho \sum_{i:d_i\leq x}d_i^3.
\end{equation}
The cumulative mass fraction is thus given by
\begin{equation}
    \label{eq:cumulative-mass-R-R}
    P(x) = \frac{M_{\leq x}}{M}.
\end{equation}

The cumulative mass fraction is often represented by using a Rosin-Ramler\ \cite{Rosin-Rammler-1933} or Swebrec model\ \cite{Ouchterlony2005,Ouchterlony2019}.  The Rosin-Rammler (R-R) model, commonly used in the mining industry and which is used in this paper, is given by
\begin{equation}
    P(x) = 1 - \exp\left\{-\left(\dfrac{x}{x_{c}}\right)^{n}\right\},
    \label{eq:Rosin-Rammler-critical-size}
\end{equation}
where $P(x)$ is the cumulative mass fraction less than size $x$, $x_{c}$ is called the critical size, and $n$ is the uniformity index\ \cite{Rosin-Rammler-1933}.  The Rosin-Ramler model is a two-parameter Weibull cumulative distribution function\ \cite{king2017}. The mean $\bar{x}$ is
\begin{equation}
	\bar{x} = x_{c}\Gamma\left(1+\dfrac{1}{n}\right),
    \label{eq:Rosin-Rammler-mean-size}
\end{equation}
where $\Gamma$ is the gamma function \cite{king2017,KuznetsovV.M.1973Tmdo}.  

It is important to note that the mean size $\bar{d}$ from \eqref{eq:average_particle_size} is not exactly the same as the mass-weighted mean particle size $\bar{x}$ in \eqref{eq:Rosin-Rammler-mean-size}.  For a given set of $N$ particles, as in \eqref{eq:average_particle_size}, the mass-weighted mean is instead given by
\begin{equation}
    \bar{x} = \frac{\sum_{i=1}^N m_id_i}{\sum_{i=1}^N m_i} = \frac{1}{M}\sum_{i=1}^N m_id_i,
\end{equation}
which provides the effective mean size of particles weighted by how much each mass contributes to the total.

One drawback to the Rosin-Rammler model \eqref{eq:Rosin-Rammler-critical-size} is the extremes---e.g., $P(x) < 15\%$ and $P(x) > 90\%$---where it often poorly reconstructs real-world data.  There are other models, such as the Swebrec model\ \cite{Ouchterlony2005}, which employs three parameters so as to better capture the extremes. However, in this paper, the focus is the relative mean particle size of different fragmented rock piles and the Rosin-Rammler model is sufficient for this purpose.

How to accurately and practically estimate $P(x)$ or $\bar{x}$ for a given pile is what constitutes the rock pile characterization problem.  Common approaches to solving this problem include using camera (or other exteroceptive) images captured of the rock pile \cite{maerz-1996,6767131}, which are subsequently processed to extract the visible part of individual particles. Another approach is to sample the rock piles directly and physically measure the sizes of the sampled particles.  The latter is usually only practical for quarries, where the materials are crushed to smaller sizes, and not in mining applications.

\subsection{QUALITATIVE VISUAL INSPECTIONS}

\begin{figure}
    \centering
    \begin{minipage}{0.47\columnwidth}
        \centering
        \subfloat[Unexcavated pile.]{\label{fig:0-150-pile-example}\includegraphics[width=\textwidth]{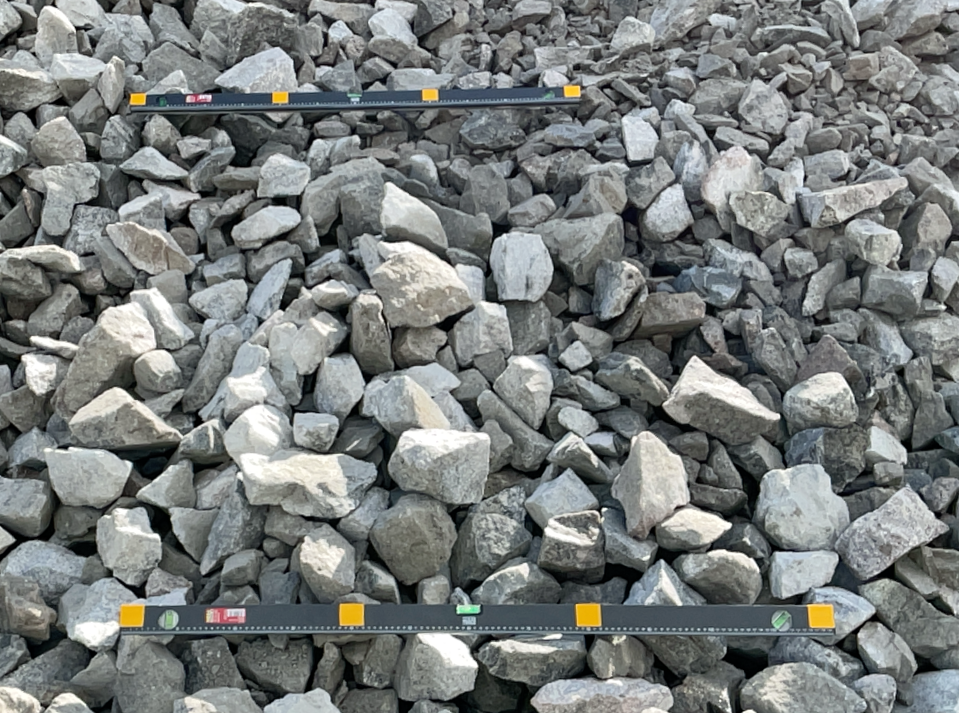}}
        
        \vspace{-0.3em} 
        
        \subfloat[Material in the bucket.]{\label{fig:0-150-bucket-example}\includegraphics[width=\textwidth]{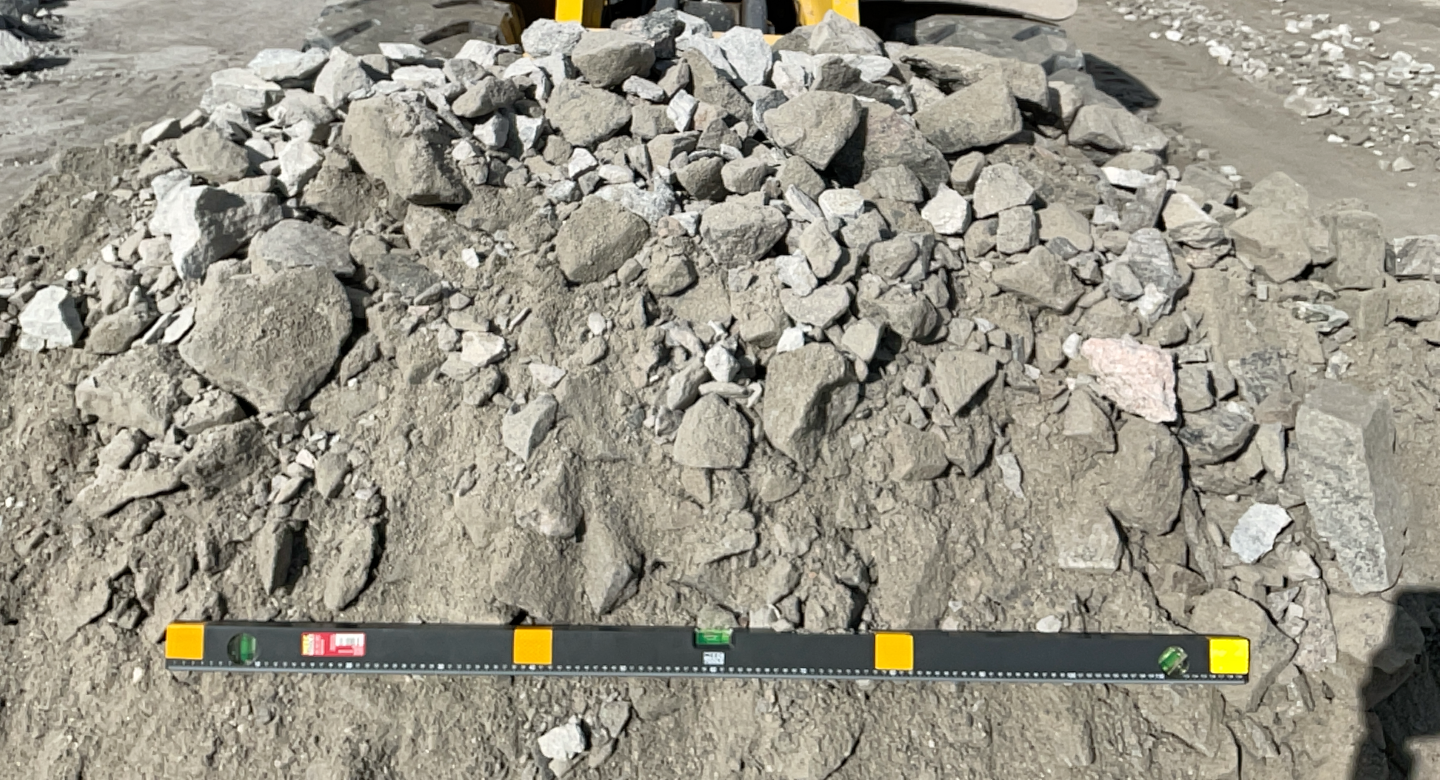}}
    \end{minipage}
    \hfill
    \begin{minipage}{0.51\columnwidth}
        \centering
        \subfloat[Material spread out on the ground.]{\label{fig:0-150-spread-example}\includegraphics[width=\textwidth]{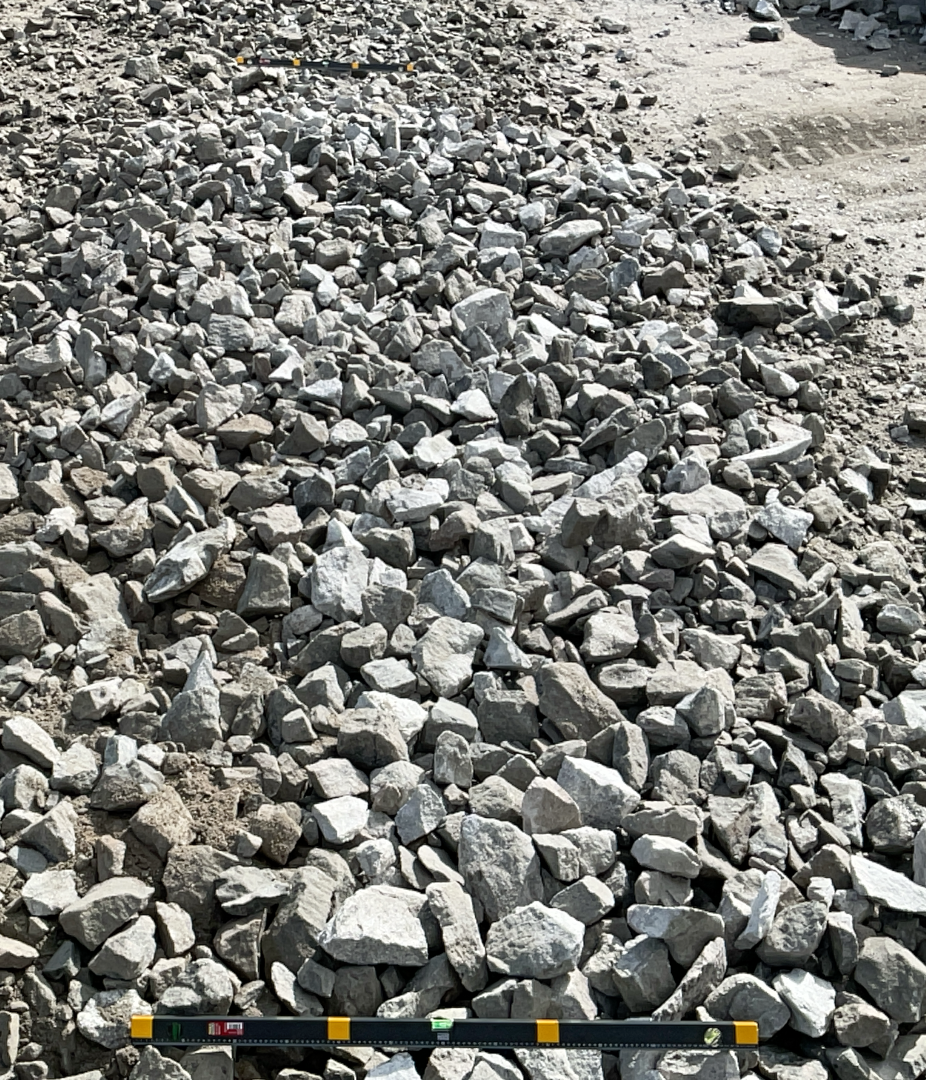}}
    \end{minipage}
    \caption{Example images at different stages of the excavation process and the variability of the visible rocks. A ruler of length 1.2 m is provided for scale and is roughly the same number of pixels in each image.}
    \label{fig:rock-image-examples}
\end{figure}

Qualitative visual inspections by equipment operators or mine personnel is a common method for assessing fragmentation in underground mining operations\ \cite{Manzoor2023}.  However, this can be inconsistent and is not repeatable because it is subjective.  Additionally, depending on the stage of the excavation process (i.e., before excavation, after excavation and the material being dumped), the appearance of a fragmented rock pile varies, as can be seen in Figure\ \ref{fig:rock-image-examples}, which highlights one of the challenges in characterizing the fragmented rock contained within the bucket of a loader after excavation.

\subsection{SIEVE ANALYSIS}
\label{sec:manual_methods}

Sieve analysis, where a ``representative'' quantity of material is passed over different sized meshes and the retained mass is used to estimate the cumulative mass distribution, is the benchmark tool for fragmentation analysis.  In the aggregate industry, sieve analysis is used to ensure products conform to standards \cite{astm-mnl32,OPSS1010,ISO-6274:1982,SS-EN-933}. 

One of the main challenges with sieve analysis is ensuring a representative sample is processed.  Given that only a relatively small amount of material can be processed (usually $\sim$30 kg), if the sample contains an unusually high amount of small or large sized rocks, the corresponding sieve size estimates will not capture the true size distribution of the rock pile.  Depending on the size range of an aggregate product, the amount of material needed to obtain a representative quantity differs, as well as the sampling frequency.  For example, the more homogeneous the aggregate product (i.e., narrow size distribution) the smaller the amount of material needed to accurately characterize the rock pile.

In mining, where the fragmented rocks are produced by drilling and blasting and lead to very heterogeneous size distributions, the application of sieve analysis is very limited.  Despite this, in the experiments described by this field report, we use sieve analysis as a comparator.

\subsection{EXTEROCEPTIVE MEASUREMENT SYSTEMS}
\label{sec:exteroceptive_methods}

\begin{figure}
  \centering
  \subfloat[Pile image 1.2 m rulers for scale.]{\label{fig:wifpfrag-image}\includegraphics[width=0.49\linewidth]{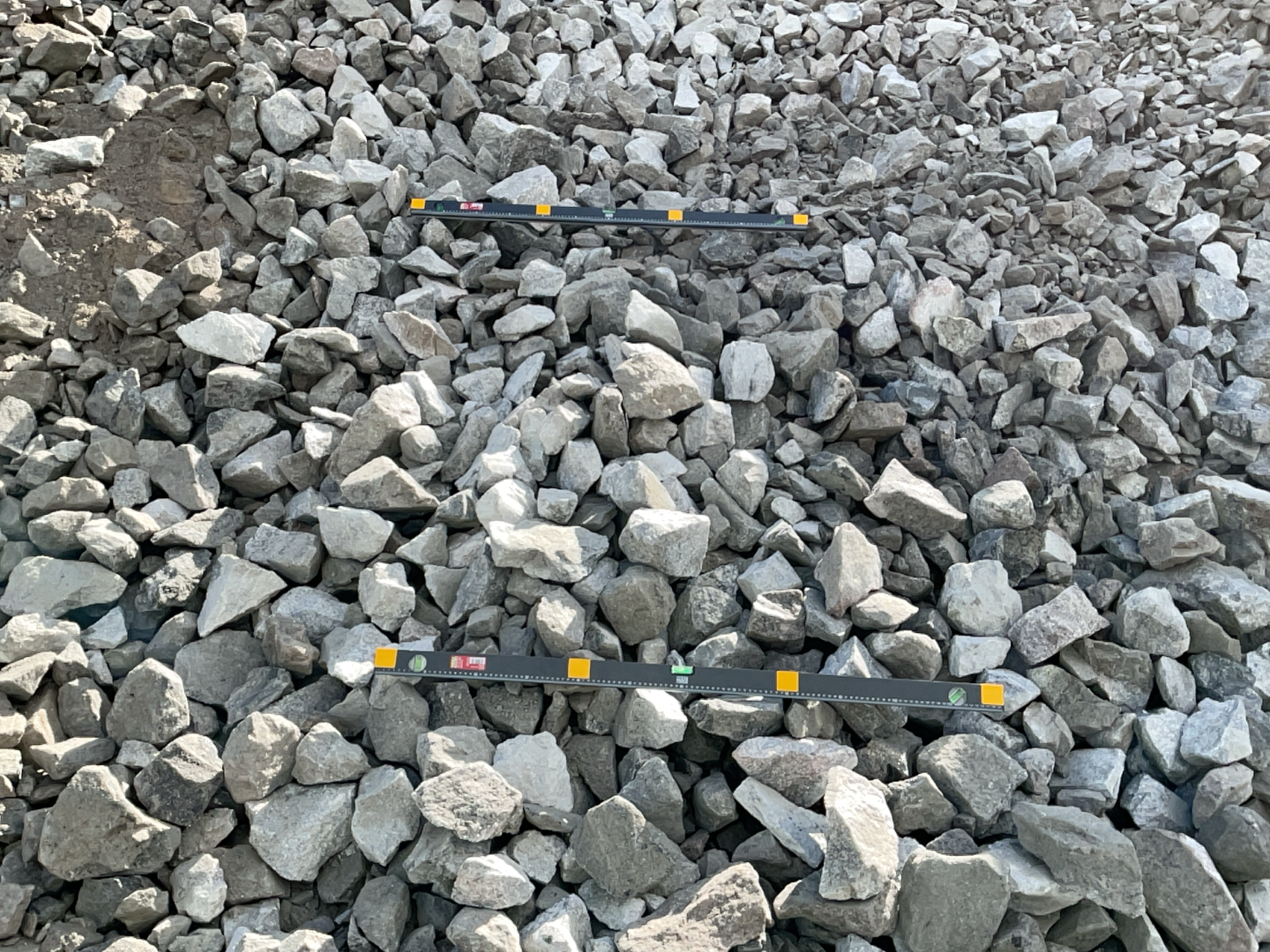}}\hfill
  \subfloat[WipFrag\texttrademark\ edge analysis network corresponding to the image (a).]{\label{fig:wifpfrag-net}\includegraphics[width=0.49\linewidth]{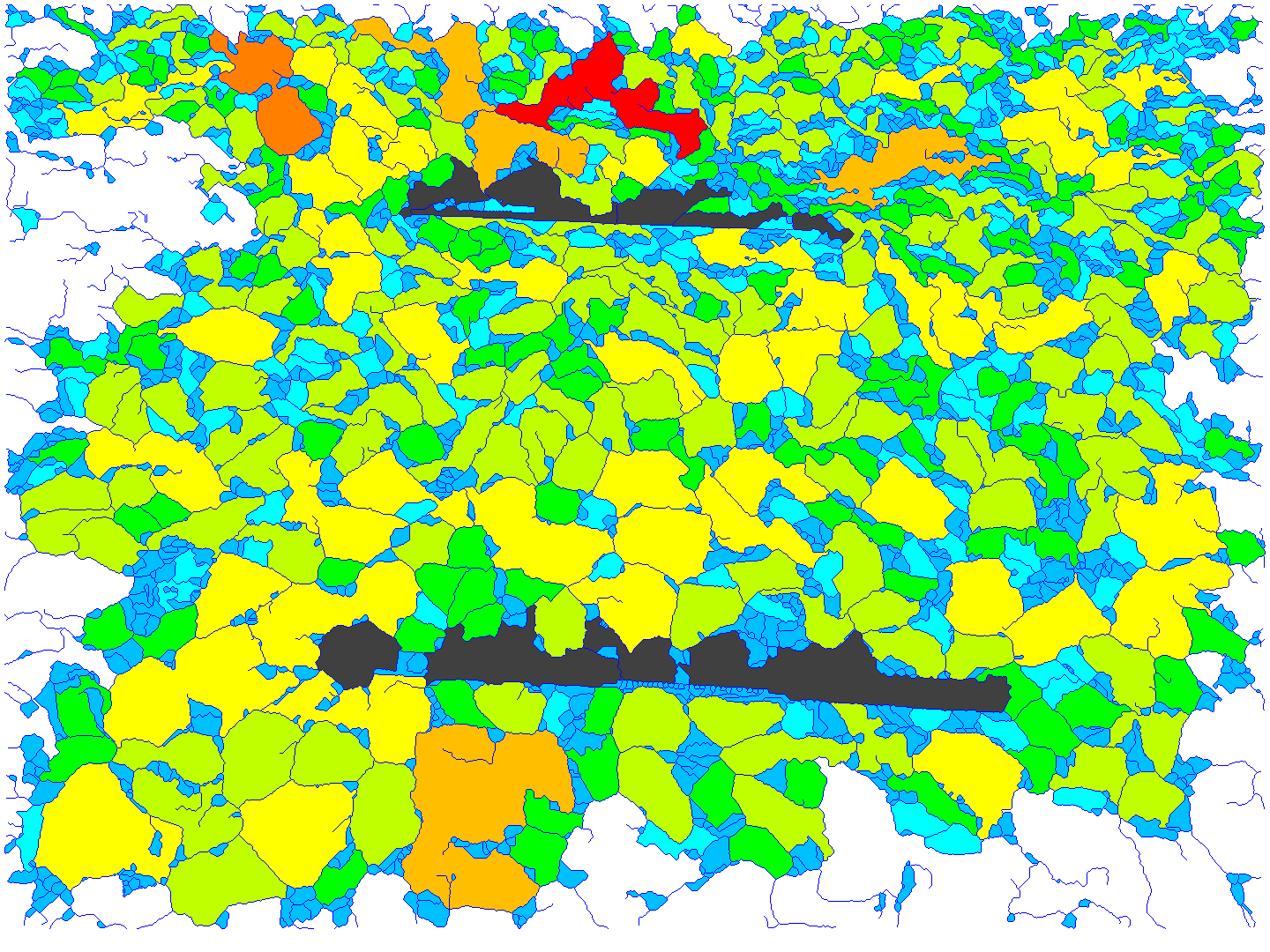}}\hfill
  \subfloat[WipFrag\texttrademark\ analysis report showing cumulative percent passing.]{\label{fig:wifpfrag-output}\includegraphics[width=\linewidth]{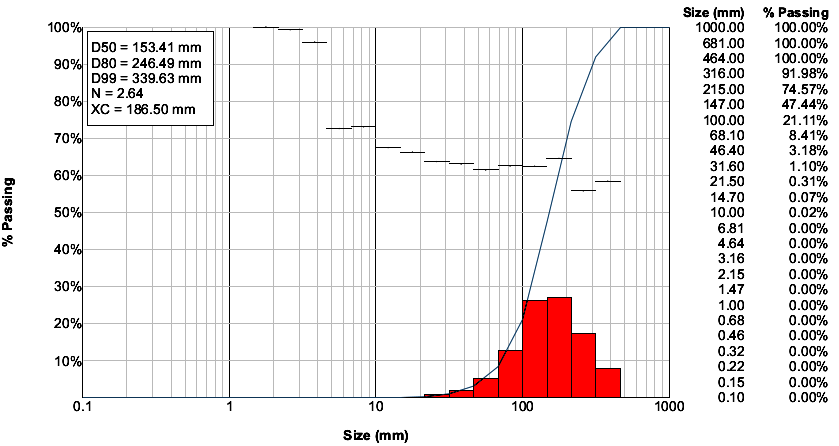}}
  \caption{Example workflow of the vision based fragmentation analysis tool WipFrag\texttrademark\ and resulting fragmentation report.}
  \label{fig:wipfrag_example}
\end{figure}

Fragmentation analysis using exteroceptive sensing is a widely used and effective method for generating particle size estimates \cite{maerz-1996,6767131,SAYADI2013318,Engin2019,sereshki2016blast}. These tools can produce size estimates quickly but they have some drawbacks and limitations \cite{Sanchidrián2009,Raina2013}. One drawback is that they only capture the surface of a pile. Heterogenous particle distributions do not settle with a consistent distribution, thus biasing surface-level (i.e., exteroceptive) techniques. One operational tweak to overcome this issue could be to acquire images/scans at a higher frequency.  For surface operations, commercial fragmentation analysis products offer offer frequent estimates, while the pile is handled, so as to avoid only observing the same surface sample \cite{Chen2023}.  Surface mining operations also have the ability to leverage advancements in Uncrewed Aerial Vehicles (UAVs) equipped with imaging hardware to obtain size estimates \cite{Bamford2017,OBOSU2025}.  An example report obtained from the commercial image based fragmentation analysis tool used in our experiments is provided in Figure\ \ref{fig:wipfrag_example}.

Recent advances in vision-based fragmentation apply image segmentation to delineate or segment the rocks within an image \cite{BAHRAINI2024108822,min14070654,BamfordThomas2021Adla}.  While these deep learning-based methods improve segmentation accuracy, they do not overcome the drawback of only capturing the surface of the pile.

\subsection{PROPRIOCEPTIVE SYSTEMS}

Proprioceptive sensors (e.g., accelerometers, gyroscopes, pressure transducers, microphones) are ideal for capturing the motions and vibrations of equipment due to collisions with excavation media. In other work, wavelet analysis demonstrated that accelerometers, mounted to a 14-tonne capacity wheel loader, could potentially be processed to effectively classify different rock piles\ \cite{ArtanU-PhD2022}. Force signals have also shown promise as a classifier input in experiments using a similar wheel loader\ \cite{FerMar-AutoCon-2020}. 

Experiments exploring the correlation between digging conditions and vibrations from an accelerometer mounted to the boom of an electric rope shovel observed differences in the signal depending on the ``digging conditions''\ \cite{KhorzoughiPhDThesis}.  In these experiments, the amplitude of a Fourier analysis of the acceleration measurements when the rope shovel operated in two different zones produced different values.  This difference was attributed to the difficulty in digging conditions between the two zones. Unfortunately, ``digging conditions'' was a qualitative metric with three categories of ``easy'', ``medium'' and ``hard'', and the relationship to size distribution was not investigated. 

Recent work\ \cite{Marshall2008, Dobson2017, AtlasCopcoAutoLoad, Fernando2019} exploits proprioceptive sensing in order to effectively modulate the bucket-rock interactions to achieve autonomous excavation using a wheel loader. The implicit connection between robotic excavation and fragmentation characteristics suggests that proprioceptive sensing for fragmentation analysis may indeed be a promising avenue of research, as illustrated by the results of this field report. 

\section{WAVELET FEATURES}
\label{sec:wavelet-rocksize-connection}

The interaction between material handling equipment and the target excavation media comprises many collisions over time.  These collisions produce a transfer of momentum (e.g., a rock dropping onto a haul truck bed or a rock entering a moving excavator bucket) that can be sensed.  The contact time of each rock is much smaller than the time to fill, for example, a truck or an excavator bucket.  Proprioceptive sensors might capture these collisions and measured signals can be analyzed further to infer information about the distribution of particle sizes.  In this work, we use wavelets to perform this analysis and study the outcome by way of full-scale experiments in representative conditions, following work started by \cite{ArtanU-PhD2022}.

For each excavation run we compute an engineered feature using the continuous wavelet transform\ \cite{Wavelets1992,Wavelets1996,Wavelets1999}, which is termed the ``wavelet feature'' $\beta$.  This wavelet feature is calculated by the maximum 
\begin{equation}
    \beta = \operatorname*{max}_{s} \frac{1}{M(\alpha_{2}-\alpha_{1})}\int_{\alpha_1}^{\alpha_2} \left(g\ast\Psi_{s}^*\right)\left(\tau\right)\text{d}\tau
    \label{eq:wavelet-feature-beta}
\end{equation}
where $\ast$ is the convolution, $M$ is the total mass of particles, $\alpha_{1}$ is the excavation start time, $\alpha_{2}$ is the excavation end time, $g$ is the input time series signal, $\Psi_{s}$ is the child wavelet that is constructed by scaling (stretching or compressing) the mother wavelet $\Psi$ by $s$ and $^{*}$ denotes the complex conjugate.  The scale $s$ can be mapped to frequency $f$ by using
\begin{equation}
	f=\frac{f_{c}}{s},
\label{eq:scale_to_frequency}
\end{equation}
where $f_{c}$ is the center frequency of the wavelet. The center frequency $f_{c}$ is the frequency maximizing the Fourier transform of the wavelet modulus to a purely periodic signal\footnote{See also https://www.mathworks.com/help/wavelet.}. Thus, the feature $\beta$ is computed by integrating over the excavation time $[\alpha_1,\alpha_2]$ and finding the frequency that yields the maximum ``response''.

In some instances during our experiments, signals from sensors such as strain gauges or pressure transducers after the collisions registered a non-zero value due to the weight of the rocks. Because wavelets are zero mean\ \cite{Wavelets1992, Wavelets1996}, it is desirable for input signals to also have zero mean. Thus, we process the signals using a high-pass filter with a cut-off frequency $f_{k}$, where frequency components greater than $f_{k}$ remain. The value of $f_{k}$ is obtained experimentally.

From experiments, the waveform not only changes in amplitude but also in shape.  Thus, an alternative wavelet feature $\zeta$ is introduced to capture these changes\ \cite{ArtMar-MFI,ArtFerMar-AIM,ArtanU-PhD2022}.  Instead of choosing the maximum value as in \eqref{eq:wavelet-feature-beta}, $\zeta$ assesses the area under the waveform and is calculated by
\begin{equation}
  \zeta = \int_{f_{\text{min}}}^{f_{\text{max}}} \frac{1}{M(\alpha_{2}-\alpha_{1})}\int_{\alpha_{1}}^{\alpha_{2}} \left(g\ast\Psi_{s}^*\right)\left(\tau\right)\text{d}\tau\text{d}f,
  \label{eq:wavelet-feature-zeta}
\end{equation}
where $f_{\text{min}} \geq f_{k}$ and $ f_{\text{max}}$ is the maximum frequency generated by wavelet analysis and is dependent on the sampling frequency.  Additionally, the complete waveform could be used as the input to classification methods.

Although these wavelet features are likely not useful for estimating the actual size or weight of particles that caused the measured interactions, when we compute the ratio between two or more wavelet features generated using the same machine but interacting with different piles of fragmented rock, insights can be gleaned about the relative characteristics of the two (different) rock piles. 

In this paper, we focus on applying the $\zeta$ wavelet feature for rock pile characterization.  The thesis\ in \cite{ArtanU-PhD2022} argues that, in fact, for two different rock piles labeled $A$ and $B$, 
\begin{equation}
    \frac{\zeta_{A}}{\zeta_{B}} \approx \frac{\bar{d}_{A}}{\bar{d}_{B}} \approx \frac{\bar{x}_{A}}{\bar{x}_{B}}.
    \label{eq:wavelet-mean-ratios}
\end{equation}
In other words, for two different rock piles $A$ and $B$, excavated using the same machine, the ratio of the wavelet features $\zeta_A$ and $\zeta_B$ is directly proportional (without scaling) to the ratio of the mean particle sizes of the respective rock piles. This relationship holds because wavelet analysis of acceleration signals captures the impulse response of bucket-rock interactions, which relates to particle size rather than mass. Therefore, the wavelet features $\zeta$ fundamentally correspond to the mean particle size $\bar{d}$. However, when comparing two material piles with comparable uniformity indices (i.e., $n_A \simeq n_B$), a useful numerical relationship emerges: $\bar{d}_A/\bar{d}_B \approx x_A/x_B$. For the crushed aggregate materials in this study, which originate from the same quarry with comparable uniformity indices, this relationship holds sufficiently well to enable direct comparison with industry-standard mass-weighted measurements.

The primary contribution of this field research paper is to test this hypothesis at scale by using commercial excavation equipment in realistic operating conditions.  We compare the wavelet feature results against both a vision-based (exteroceptive) system as well as sieve analysis.

\section{EXPERIMENTAL SETUP}
\label{sec:experimental-setup}

The remaining sections of this field report introduce our experimental setup, including the equipment used and operating environment, followed by a detailed presentation of the experimental results.  These experiments involved the use of full-scale commercial equipment and we directly compare our results with two alternative approaches---sieve analysis of sampled materials as well as a reputable and widely-used vision-based technology---as ground truths.

\subsection{EXCAVATION EQUIPMENT AND TEST OPERATORS}

The experiments presented in this paper were all done using a 2025 model year Epiroc Scooptram ST18 SG battery-electric load-haul-dump (LHD) machine, as shown in Figure\ \ref{fig:material-handling-example}.  The LHD was equipped with a bucket having a bulk volume capacity of $6.2\ \text{m}^3$.  

Excavation trials were performed by two operators, labeled Operator A and Operator B, and allowed for the evaluation of robustness and generalizability of the automated fragmentation estimation methods with varying excavation styles. Operator A was significantly more experienced than Operator B.  The two operators were instructed to obtain a full bucket of material and to not excessively scrape the ground. Otherwise, the operators both had full control of the excavation process during all trials. 

\subsection{EXCAVATION MATERIALS}

\begin{figure}
  \centering
  \subfloat[0/32 pile.]{\label{fig:0-32-pile}\includegraphics[width=0.49\linewidth]{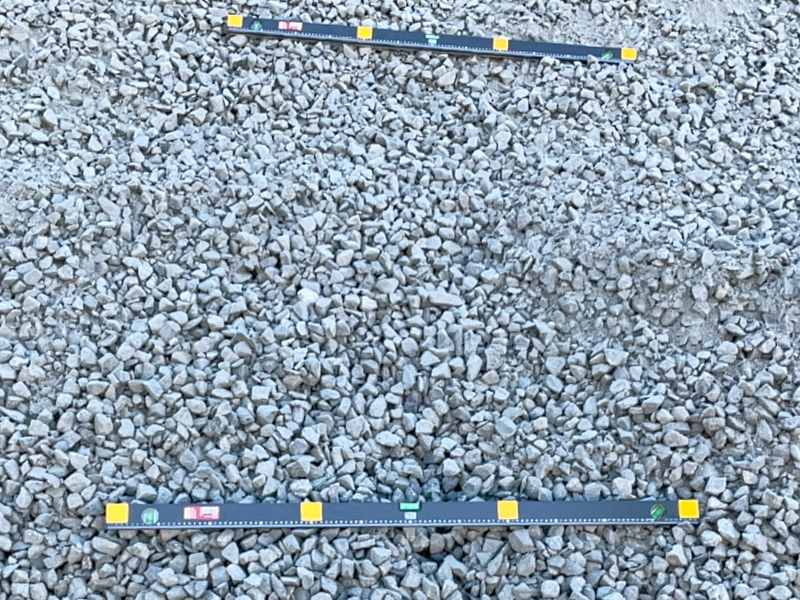}}\hspace{1 pt}
  \subfloat[0/63 pile.]{\label{fig:0-63-pile}\includegraphics[width=0.49\linewidth]{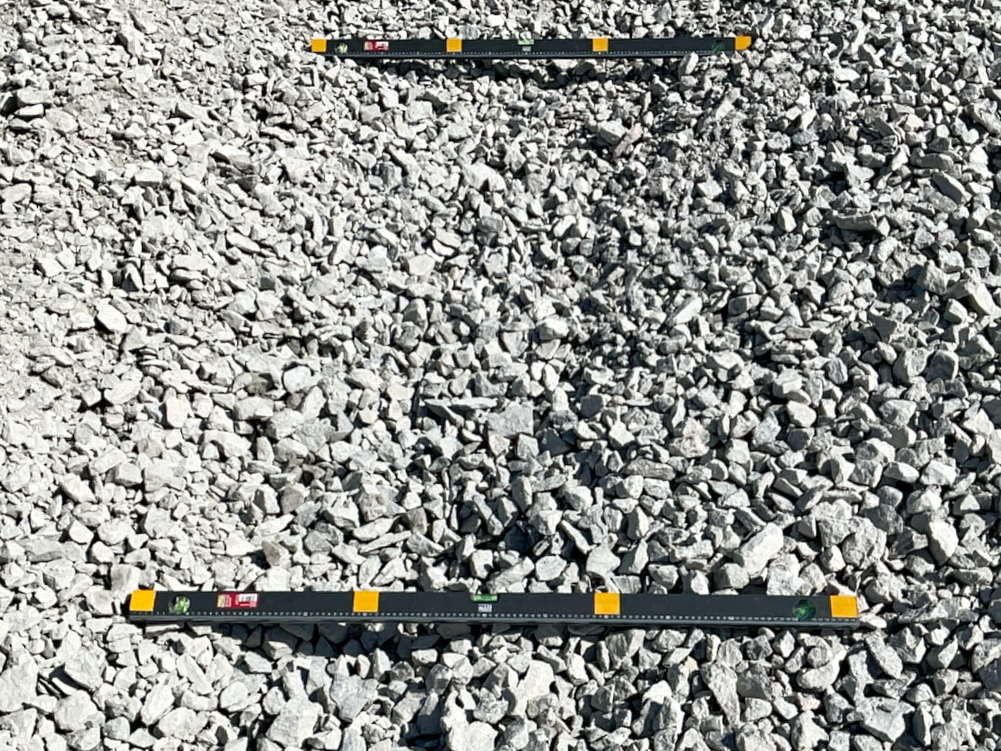}}\hspace{1 pt}
  \subfloat[0/90 pile.]{\label{fig:0-90-pile}\includegraphics[width=0.49\linewidth]{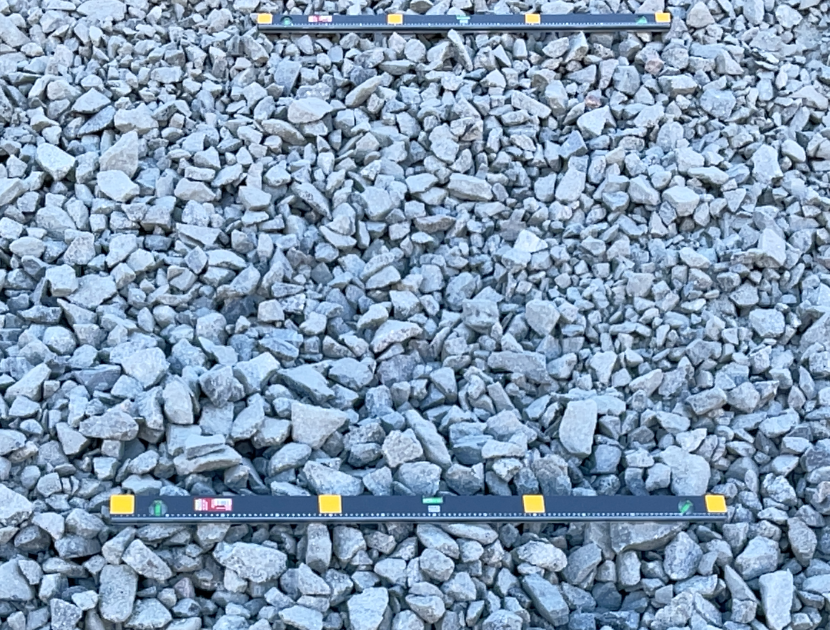}}\hspace{1 pt}
  \subfloat[0/150 pile.]{\label{fig:0-150-pile}\includegraphics[width=0.49\linewidth]{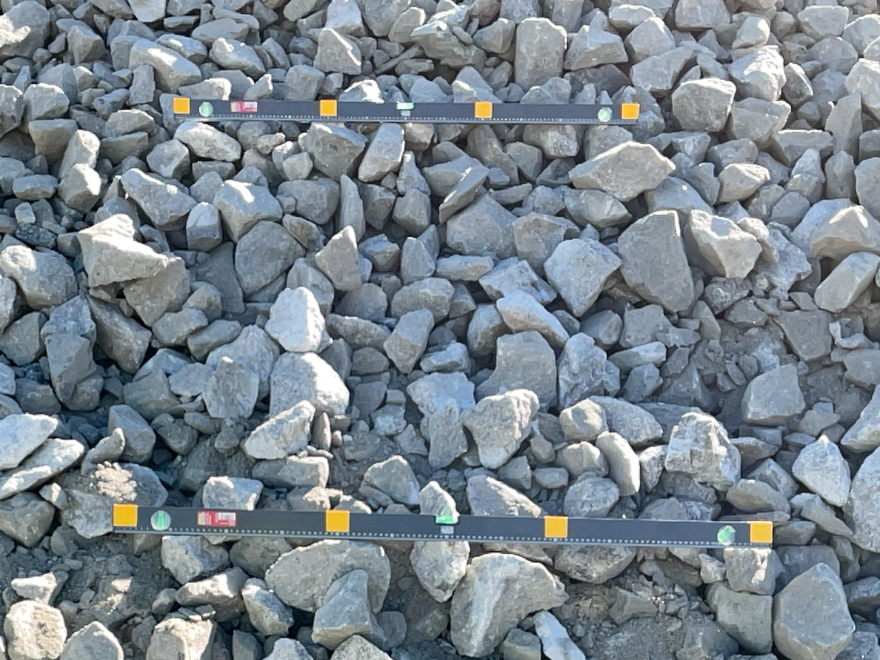}}
  \caption{Example pile images of the four semi-homogeneous aggregate piles used for excavation experiments.  Two 1.2-m wide rulers with 2-cm yellow squares spaced 0.4 m apart is provided for scale.}
  \label{fig:semi-homogeneous-pile-images}
\end{figure}

\begin{figure}
  \centering
  \subfloat[Pile image of 0/1500 trial 3.]{\label{fig:0-1500-A}\includegraphics[width=0.49\linewidth]{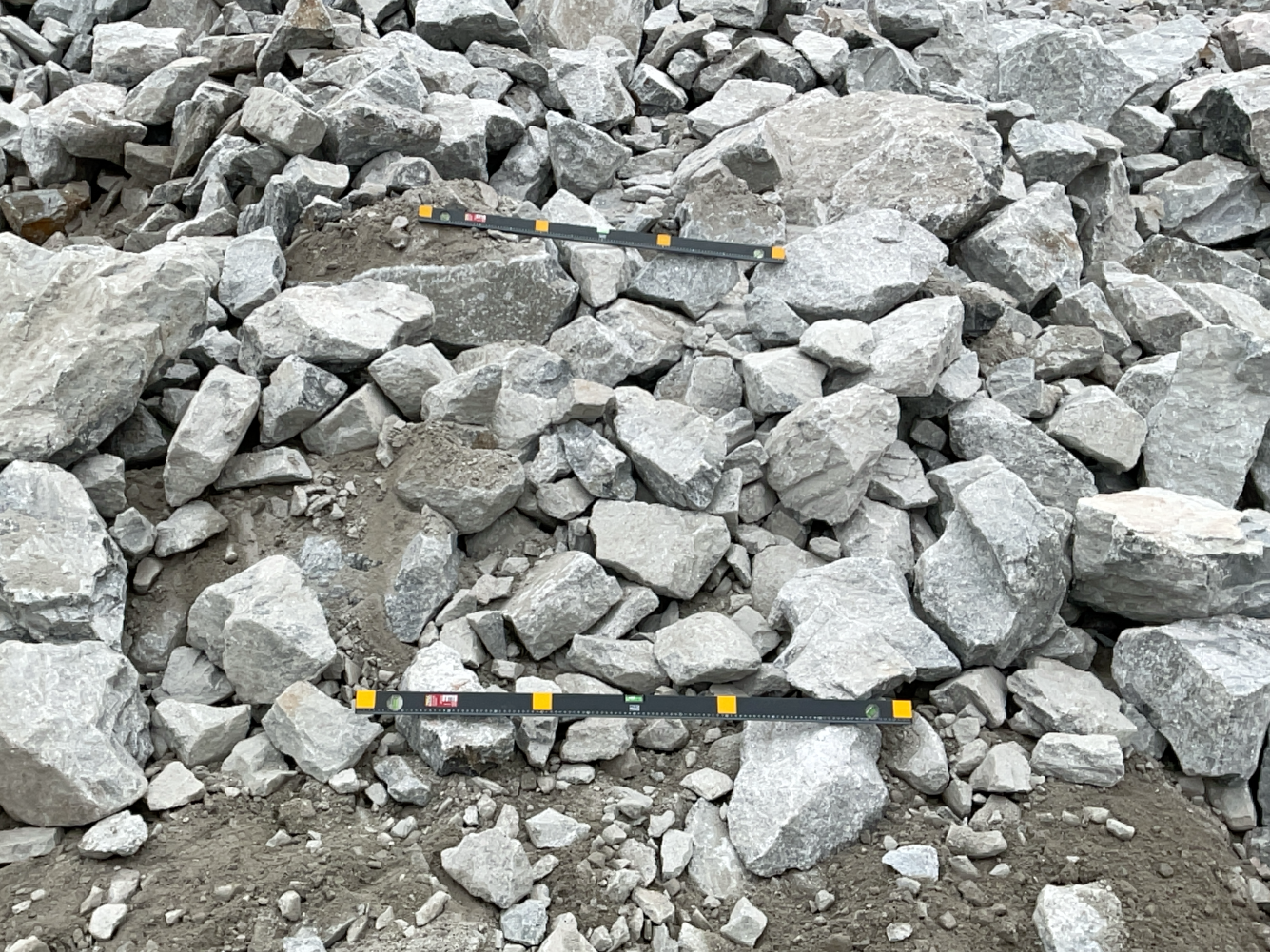}}\hspace{1 pt}
  \subfloat[Pile image of 0/1500 trial 10.]{\label{fig:0-1500-B}\includegraphics[width=0.49\linewidth]{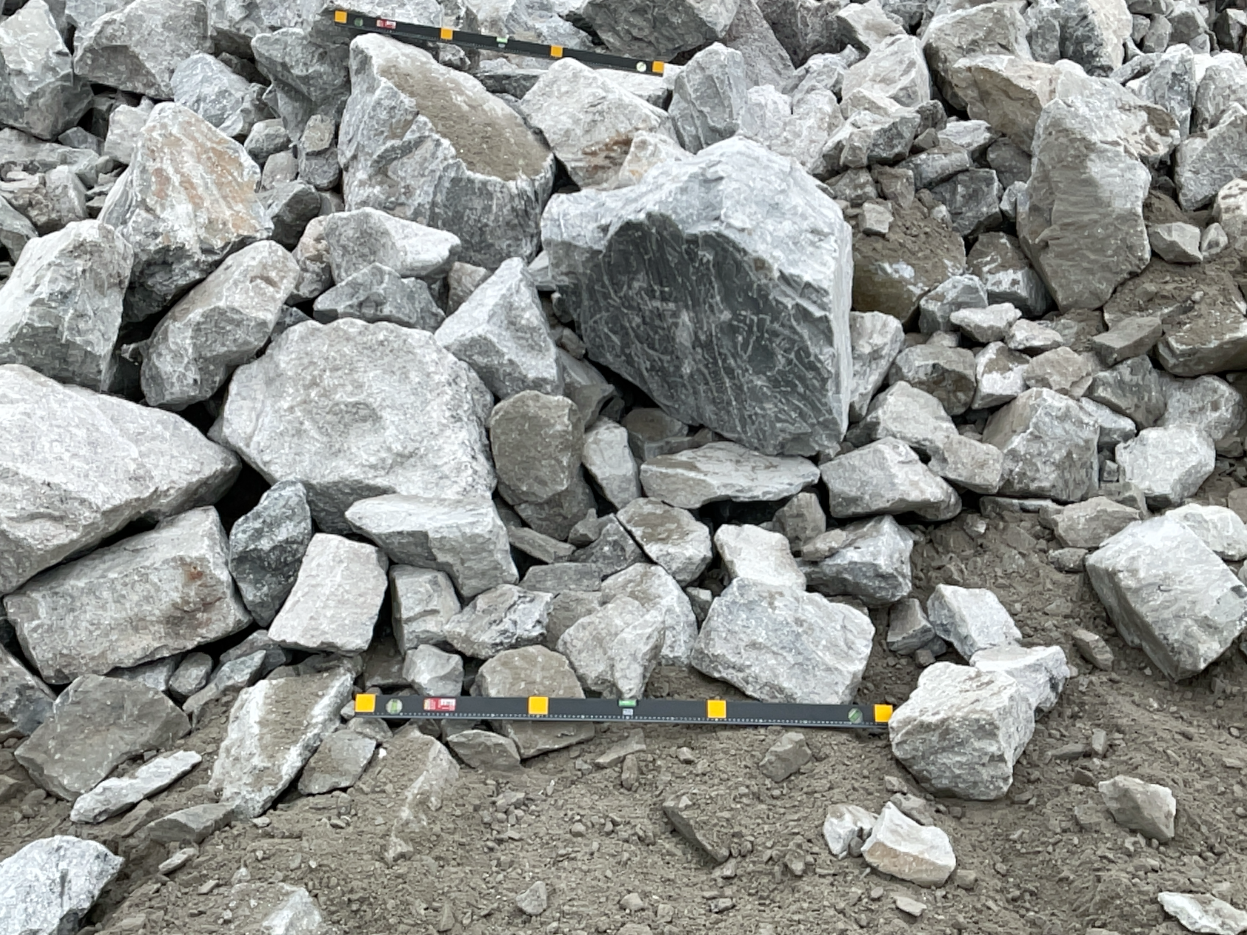}}\hspace{1 pt}
  \subfloat[Pile image of 0/1500 trial 15.]{\label{fig:0-1500-C}\includegraphics[width=0.49\linewidth]{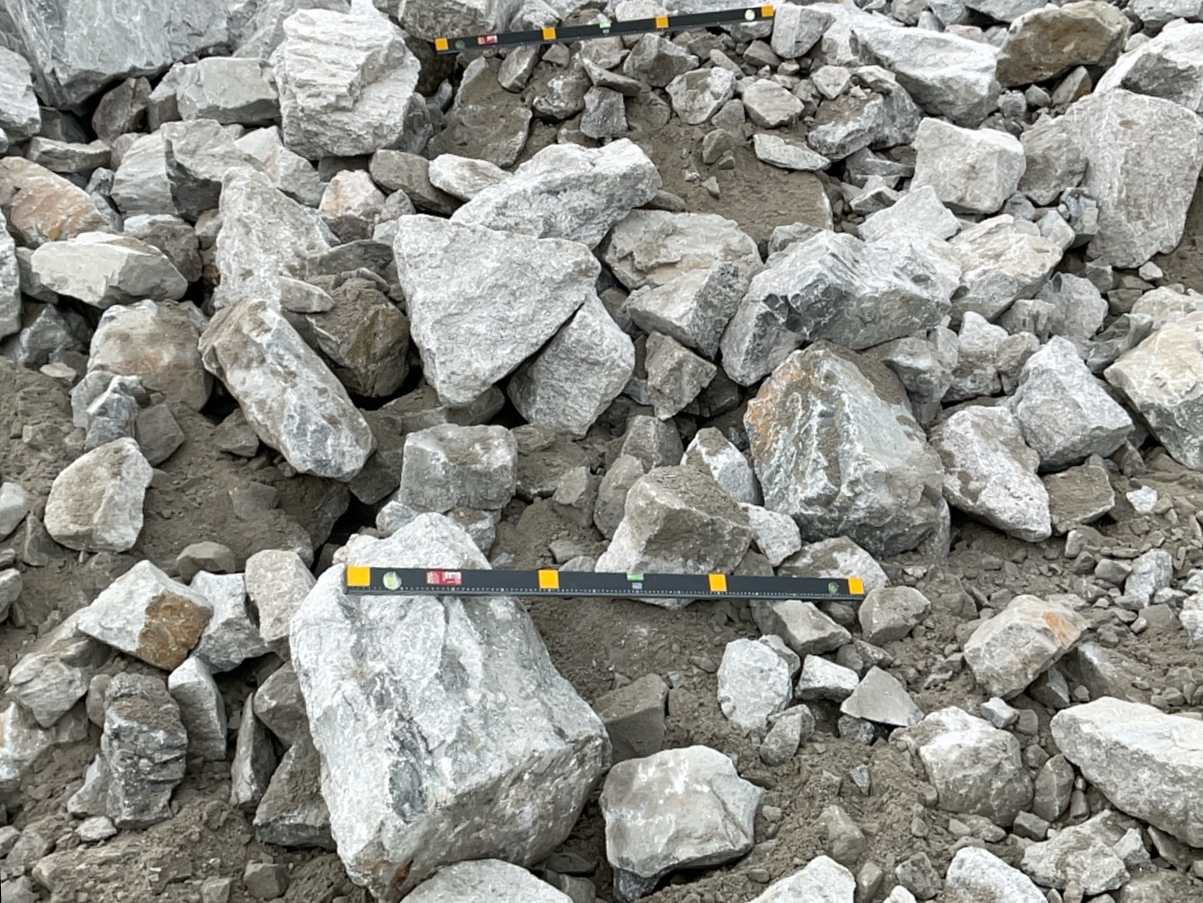}}\hspace{1 pt}
  \subfloat[Pile image of 0/1500 trial 17.]{\label{fig:0-1500-D}\includegraphics[width=0.49\linewidth]{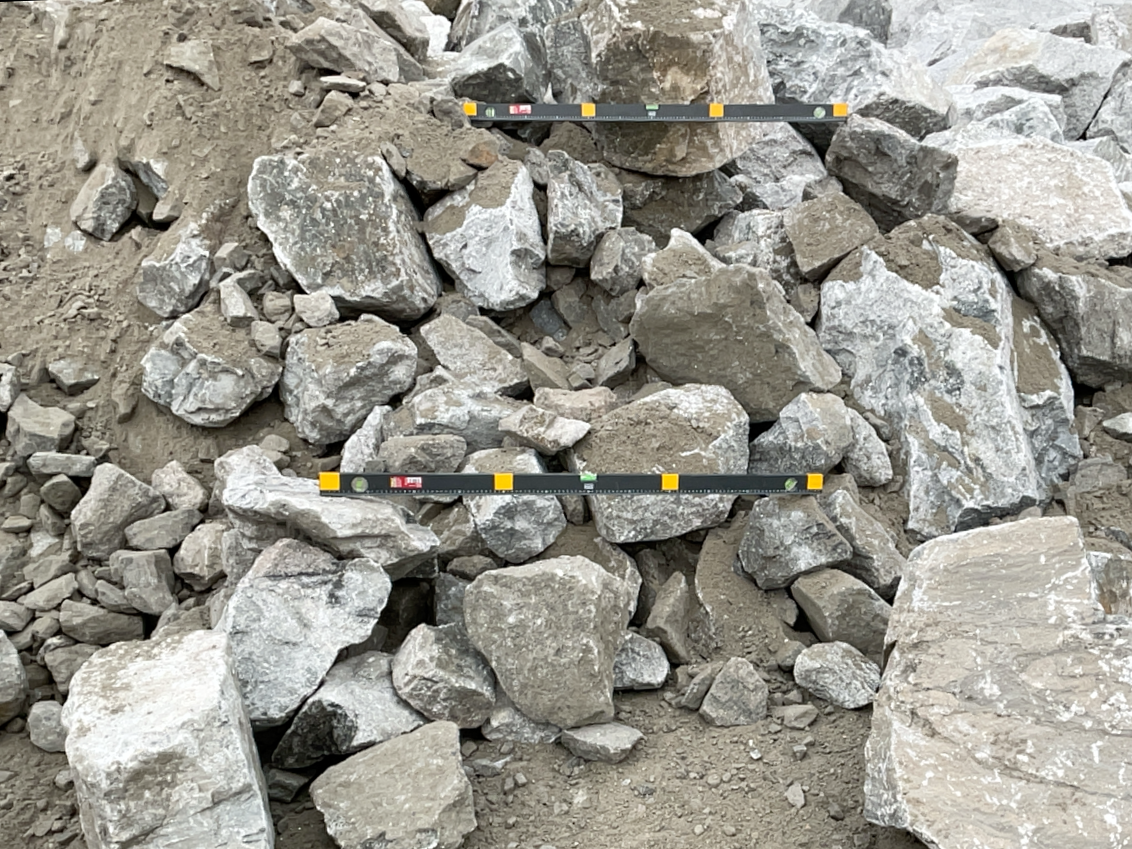}} 
  \caption{Pile images of the heterogeneous aggregate pile 0/1500.  Two 1.2-m wide rulers with 2-cm yellow squares spaced 0.4 m apart is provided for scale.}
  \label{fig:heterogeneous-pile-images}
\end{figure}

All of the experiments presented in this field report were performed at an aggregate quarry located near Eker, Sweden, that produces crushed aggregate with different size distributions for specific industrial purposes.  Five rock pile types were used in this study.  Four of these piles were generated by using crushed materials that were labeled 0/32, 0/63, 0/90 and 0/150, where the naming convention $x_{\text{min}}/x_{95}$ indicates that particles range from approximately $x_{\text{min}}$ mm to a size where 95\% of particles by mass are smaller than $x_{95}$ mm. These piles were semi-homogeneous in rock size and an example image of each pile is shown in Figure\ \ref{fig:semi-homogeneous-pile-images}. The fifth rock pile, labeled 0/1500, was post blast fragmented rocks with fragments larger than 1.5 m removed.  Figure\ \ref{fig:heterogeneous-pile-images} illustrates some of the variability in rock sizes of the 0/1500 rock pile, which was the largest. By comparison with photos of blasted rock fragments from underground mining operations \cite{Manzoor2023} and through discussions with mine operators, the five piles adequately span the range of size distributions that could be encountered in practice. 

%
%
%
%

\subsubsection{SIEVE ANALYSIS}
\label{sec:sieve_analysis}
\begin{figure}
  \centering
  \includegraphics[width=\linewidth]{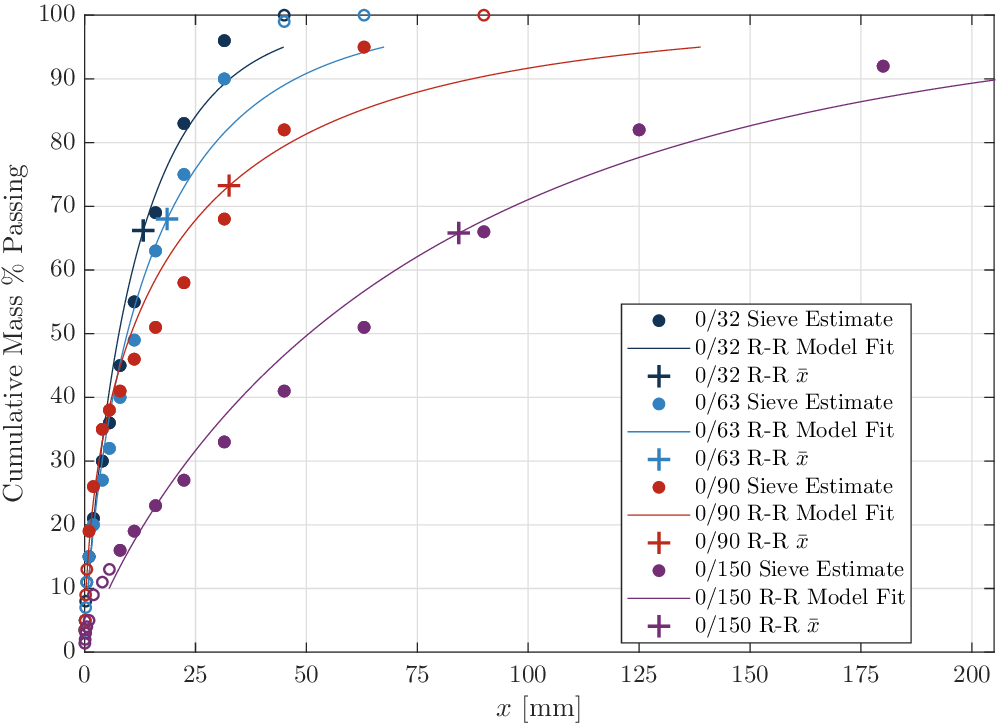}
  \caption{Laboratory sieve estimates of the cumulative mass percentage and corresponding Rosin-Rammler estimate for the four semi-homogeneous rock piles. Circles indicate all sieve data points, while filled circles show data points used in the regression (constrained to $P(x_{\text{min}}) \geq 15\%$ and $P(x_{\text{max}})$ in the range 90--96\%). Cross markers ($\times$) indicate the mean particle size $\bar{x}$ for each pile.}
  \label{fig:semi-homogeneous-size-estimates}
\end{figure}

The four semi-homogeneous rock piles were analyzed by a Swedish accredited laboratory to estimate the particle size distribution by sieving to standard SS-EN 933-1 \cite{SS-EN-933} and material density to standard SS-EN 1097-6 \cite{SS-EN-1097-6:2013}.  The laboratory estimates of cumulative mass percentage passing from the acquired samples are provided in \cref{tbl:svevia-sieve-results} and illustrated in Figure\ \ref{fig:semi-homogeneous-size-estimates}. The density of rock was 2.63\ $\rm t/m^{3}$.

\begin{table*}
\centering
\caption{Sieve estimates for the four semi-homogeneous rock piles, based on a sample of 25 kg for pile 0/32, 70 kg for pile 0/63, 90 kg for pile 0/90 and 225 kg for pile 0/150, obtained from an accredited laboratory. Shaded cells indicate data points used in the Rosin-Rammler regression (constrained to $P(x_{\text{min}}) \geq 15\%$ and $P(x_{\text{max}})$ in range 90--96\%).}
\label{tbl:svevia-sieve-results}
\begin{tabular}{lllllllllllllllllll}
\toprule
\multicolumn{1}{c}{}         & \multicolumn{18}{c}{\textbf{Cumulative Mass \% Passing}} \\ \midrule
\multicolumn{1}{l|}{\textbf{Pile 0/32}}         & \multicolumn{1}{c}{3.5}            & \multicolumn{1}{c}{5}              & \multicolumn{1}{c}{8}             & \multicolumn{1}{c}{11}           & \multicolumn{1}{c}{\cellcolor{lightgray}15}         & \multicolumn{1}{c}{\cellcolor{lightgray}21}         & \multicolumn{1}{c}{\cellcolor{lightgray}30}         & \multicolumn{1}{c}{\cellcolor{lightgray}36}           & \multicolumn{1}{c}{\cellcolor{lightgray}45}         & \multicolumn{1}{c}{\cellcolor{lightgray}55}            & \multicolumn{1}{c}{\cellcolor{lightgray}69}          & \multicolumn{1}{c}{\cellcolor{lightgray}83}            & \multicolumn{1}{c}{\cellcolor{lightgray}96}            & \multicolumn{1}{c}{100}         & \multicolumn{1}{c}{}   & \multicolumn{1}{c}{}   & \multicolumn{1}{c}{}    & \multicolumn{1}{c}{}    \\ 
\multicolumn{1}{l|}{\textbf{Pile 0/63}}         & \multicolumn{1}{c}{3.4}            & \multicolumn{1}{c}{5}              & \multicolumn{1}{c}{7}             & \multicolumn{1}{c}{11}           & \multicolumn{1}{c}{\cellcolor{lightgray}15}         & \multicolumn{1}{c}{\cellcolor{lightgray}20}         & \multicolumn{1}{c}{\cellcolor{lightgray}27}         & \multicolumn{1}{c}{\cellcolor{lightgray}32}           & \multicolumn{1}{c}{\cellcolor{lightgray}40}         & \multicolumn{1}{c}{\cellcolor{lightgray}49}          & \multicolumn{1}{c}{\cellcolor{lightgray}63}          & \multicolumn{1}{c}{\cellcolor{lightgray}75}            & \multicolumn{1}{c}{\cellcolor{lightgray}90}            & \multicolumn{1}{c}{99}          & \multicolumn{1}{c}{100}         & \multicolumn{1}{c}{}            & \multicolumn{1}{c}{}             & \multicolumn{1}{c}{}             \\ 
\multicolumn{1}{l|}{\textbf{Pile 0/90}}         & \multicolumn{1}{c}{3.5}            & \multicolumn{1}{c}{5}              & \multicolumn{1}{c}{9}             & \multicolumn{1}{c}{13}           & \multicolumn{1}{c}{\cellcolor{lightgray}19}         & \multicolumn{1}{c}{\cellcolor{lightgray}26}         & \multicolumn{1}{c}{\cellcolor{lightgray}35}         & \multicolumn{1}{c}{\cellcolor{lightgray}38}           & \multicolumn{1}{c}{\cellcolor{lightgray}41}         & \multicolumn{1}{c}{\cellcolor{lightgray}46}            & \multicolumn{1}{c}{\cellcolor{lightgray}51}          & \multicolumn{1}{c}{\cellcolor{lightgray}58}            & \multicolumn{1}{c}{\cellcolor{lightgray}68}            & \multicolumn{1}{c}{\cellcolor{lightgray}82}          & \multicolumn{1}{c}{\cellcolor{lightgray}95}          & \multicolumn{1}{c}{100}         & \multicolumn{1}{c}{}             & \multicolumn{1}{c}{}             \\ 
\multicolumn{1}{l|}{\textbf{Pile 0/150}}        & \multicolumn{1}{c}{1.4}            & \multicolumn{1}{c}{2}              & \multicolumn{1}{c}{3}             & \multicolumn{1}{c}{4}            & \multicolumn{1}{c}{5}          & \multicolumn{1}{c}{9}          & \multicolumn{1}{c}{11}         & \multicolumn{1}{c}{13}           & \multicolumn{1}{c}{\cellcolor{lightgray}16}         & \multicolumn{1}{c}{\cellcolor{lightgray}19}            & \multicolumn{1}{c}{\cellcolor{lightgray}23}          & \multicolumn{1}{c}{\cellcolor{lightgray}27}            & \multicolumn{1}{c}{\cellcolor{lightgray}33}            & \multicolumn{1}{c}{\cellcolor{lightgray}41}          & \multicolumn{1}{c}{\cellcolor{lightgray}51}          & \multicolumn{1}{c}{\cellcolor{lightgray}66}         & \multicolumn{1}{c}{\cellcolor{lightgray}82}             & \multicolumn{1}{c}{\cellcolor{lightgray}92}             \\ \midrule
\multicolumn{1}{l|}{\textbf{Sieve (mm)}}        & \multicolumn{1}{c}{0.063}          & \multicolumn{1}{c}{0.125}           & \multicolumn{1}{c}{0.25}           & \multicolumn{1}{c}{0.5}            & \multicolumn{1}{c}{1}          & \multicolumn{1}{c}{2}          & \multicolumn{1}{c}{4}        & \multicolumn{1}{c}{5.6}            & \multicolumn{1}{c}{8}        & \multicolumn{1}{c}{11.2}             & \multicolumn{1}{c}{16}         & \multicolumn{1}{c}{22.4}           & \multicolumn{1}{c}{31.5}             & \multicolumn{1}{c}{45}           & \multicolumn{1}{c}{63}           & \multicolumn{1}{c}{90}         & \multicolumn{1}{c}{125}          & \multicolumn{1}{c}{180}          \\ \bottomrule
\end{tabular}
\end{table*}

Sieve data was regressed to estimate the Rosin-Rammler model parameters $n$ and $x_{c}$. As noted in Section~\ref{sec:fragmented-rock}-\ref{sec:rock_pile_models}, the Rosin-Rammler model poorly fits distribution extremes. Additionally, sieve sampling is asymmetric: \cref{tbl:svevia-sieve-results} shows 4--8 densely-spaced measurements below $P(x) = 15\%$ but only 1--2 sparse measurements above $P(x) = 90\%$.
To address these issues, sieve estimates were constrained to $P(x_{\text{min}}) \geq 15\%$ (corresponding to $x_{\text{min}} \approx 1$~mm) and $P(x_{\text{max}})$ in the range 90--96\%.

The regressed Rosin-Rammler models from the sieve estimates for the four piles 0/32, 0/63, 0/90 and 0/150 are shown in Figure\ \ref{fig:semi-homogeneous-size-estimates} and model parameters provided in \cref{tbl:sieve-R-R-fit}. The data points and corresponding model fits shown in Figure\ \ref{fig:semi-homogeneous-size-estimates} are just one estimate of the the four rock piles' size distribution dependent on a single sample.  

\begin{table}
    \caption{Regressed Rosin-Rammler model parameters derived from the sieve estimates in \cref{tbl:svevia-sieve-results}. Regression constrained to $P(x_{\text{min}}) \geq 0.15$ and $P(x_{\text{max}}) \leq 0.96$ to minimize bias from asymmetric sieve sampling density.}

    \label{tbl:sieve-R-R-fit}
    \centering
    \begin{tabular}{l|cccc}
        \toprule
        \bf Parameter & \bf 0/32 & \bf 0/63 & \bf 0/90 & \bf 0/150 \\
        \midrule
        $n$ & 0.8322 & 0.7506 & 0.5664 & 0.8519 \\
        $x_{c}$ [mm] & 12 & 16 & 20 & 78 \\
        $\bar{x}$ [mm] & 13 & 19 & 33 & 84 \\
        \bottomrule
    \end{tabular}
\end{table}

\subsubsection{VISION-BASED FRAGMENTATION ANALAYSIS}

All five rock pile types were also analyzed by using the commercial-grade vision-based fragmentation analysis tool called WipFrag\texttrademark\ (v4.0.30.0).  WipFrag\texttrademark\ is industry-leading software designed specifically for this purpose and widely employed by mining companies around the world.

For each fragmentation analysis reported in this paper, a raw image was taken (using an iPad Air running iOS 18.5) that includes up to two rulers, as shown in the sample Figure\ \ref{fig:wifpfrag-image}.  After manually providing the scale to WipFrag\texttrademark\ by indicating the lengths of the rulers in the frame and eliminating irrelevant parts of the image by selecting appropriate pixels in the image, WipFrag\texttrademark's \textit{Deep Learning} edge detection feature was used to segment the image into ``particles'', as shown by the example in Figure\ \ref{fig:wifpfrag-net}.  The WipFrag\texttrademark\ software uses these boundaries to estimate the distribution of particle sizes in the rock pile and, in turn, model parameters such as those shown by the example output of Figure\ \ref{fig:wifpfrag-output}.  

\subsection{EXCAVATION DATA SET}

An integrated data logging system based on the Robot Operating System 2 (ROS 2) was developed to capture various signals with each signal source described below.  Additionally, images of the rock piles at various stages of excavation were captured and used as input to the vision-based fragmentation analysis tool WipFrag\texttrademark.

\subsubsection{BUCKET AND BOOM IMUS}

\begin{figure}
  \centering
  \includegraphics[width=\linewidth]{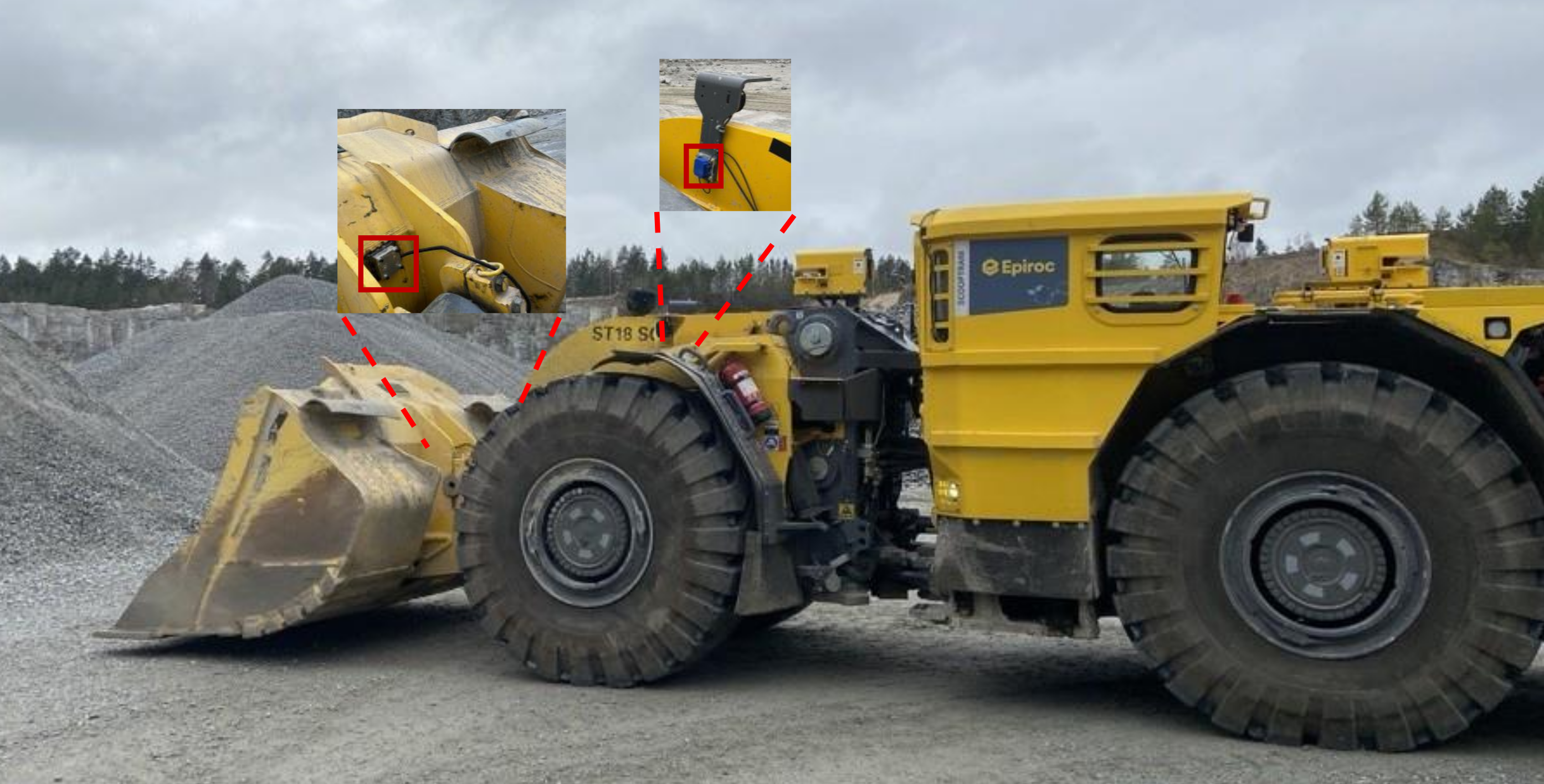}
  \caption{Locations (highlighted with a red box) of the Xsens MTi-30 IMUs installed on the Epiroc Scooptram ST18 SG LHD.  The bucket mounted unit was enclosed with an aluminum structure to minimize potential damage from rock falls.}
  \label{fig:st18-imu-locations}
\end{figure}

XSens MTi-30-2A8G4 Inertial Measurement Units (IMUs) were installed on the LHD, one on the bucket and one on the boom arm, as shown in Figure\ \ref{fig:st18-imu-locations}. IMU signals were sampled at 1000~Hz.  During the trials, an issue occurred with the bucket mounted IMU after the first day.  It was replaced with another IMU of the same model and type.  Therefore, our results from the bucket mounted IMU are separated into \emph{IMU 1} for to Day 1 trials, and \emph{IMU 2} for Days 2 and 3.

\subsubsection{LIFT CYLINDER FORCE}

Pressure readings from one of the two lift cylinders were obtained using Gefran PT-400-SEG14-A-ZVG/DE pressure transducers with a measurement range of 0--400~bar, 4--20~mA output signal, and manufacturer-specified accuracy of $\pm$0.25\% full scale ($\pm$1~bar), sampled at a rate of 250~Hz. The lift cylinders have base and rod cross sections of $A_{\text{base}} = 0.031415\ \rm m^{2}$ and $A_{\text{rod}} = 0.019175\ \rm m^{2}$, respectively.  The lift cylinder force was then calculated by
\begin{equation}
    F = 2 \times \left( A_{\text{base}} P_{\text{base}} - A_{\text{rod}} P_{\text{rod}}\right).
    \label{eq:cylinder-force}
\end{equation}
Pressure signals $P_{\text{base}}$ and $P_{\text{rod}}$ were acquired and published using a custom signal acquisition setup based on Arduino architecture hardware.

\subsubsection{CYLINDER EXTENSIONS}

Cylinder extension distances were obtained from in-cylinder mounted Waycon MH-C-0790M-N16G-3-A01 magnetostrictive position sensors with manufacturer-specified linearity of $\pm$0.05\%. The bucket cylinder has a stroke length of 550 mm and the cylinder extension is denoted by $D_{\text{bucket}}$.  The lift cylinder had a stroke length of 789 mm and the cylinder extension is denoted by $D_{\text{lift}}$. Position signals were collected using the same Arduino-based signal acquisition setup as used for pressures.

\subsubsection{MACHINE SPEED}

Speed estimates were obtained through the machine's proprietary CAN bus setup at a rate of 20~Hz.

\section{FIELD EXPERIMENT RESULTS}
\label{sec:results}

Over a period of three days, a total of 225 excavation trials were performed with the two operators at an operating aggregate quarry near Eker, Sweden.  \cref{tbl:data_set_excavation} provides an overview of the number of excavation trials performed by each operator, in each pile, and the number of image sets captured for vision-based fragmentation analysis.

\begin{table}
	\caption{Excavation experiment trial numbers.}
	\label{tbl:data_set_excavation}
	\begin{center}
    \begin{tabular}{l|ccc}
    \toprule
    \bf Pile & \bf  Operator A & \bf Operator B & \bf No.\ of Image Sets \\
    \midrule
    0/32   & 40 & 10 & 10 \\ 
    0/63   & 30 & 10 & 10 \\ 
    0/90   & 40 & 10 & 10 \\ 
    0/150  & 40 & 10 & 15 \\ 
    0/1500 & 30 & 5 & 35 \\       
    \bottomrule
    \end{tabular}
	\end{center}
\end{table}

The MATLAB\textregistered\ numerical computing software (version 2024b) was used to compute the wavelet features described in \cref{sec:wavelet-rocksize-connection}. In particular, the built-in function \texttt{cwt}\footnote{See also \url{https://www.mathworks.com/help/wavelet}.} in MATLAB's Wavelet Toolbox\texttrademark\ (version 24.4) was used to implement the 1-D continuous wavelet transform and then utilized for processing acceleration and force signals to generate the wavelet feature, $\zeta$.  Default \texttt{cwt} parameters were used.  

\subsection{EXCAVATION WINDOW AND OPERATOR VARIABILITY}

An excavation time window $t\in[\alpha_1,\alpha_2]$ is needed to estimate the wavelet feature $\zeta$ from \eqref{eq:wavelet-feature-zeta}. The start time $\alpha_1$, where the bucket begins to interact with the rock pile, was chosen as the time when the derivative of the bucket IMU acceleration $\lVert{\dot{a}_{z}\rVert}$ exceeded a threshold, $\gamma_{\text{bucket}} > 750~{\rm m}/{\rm s}^3$ when analyzing the bucket IMU acceleration signal and the boom IMU acceleration $\lVert{\dot{a}_{x} \rVert}$, exceeded a threshold, $\gamma_{\text{boom}} > 500~{\rm m}/{\rm s}^3$ when analyzing the boom IMU acceleration signal. The $z$ and $x$ axis accelerations of the bucket and boom IMUs, respectively, were aligned with the direction of travel. These thresholds $\gamma_{\text{bucket}}$ and $\gamma_{\text{boom}}$ were selected through the inspection of data from multiple trials.  The end time $\alpha_{2}$ was determined by using the bucket cylinder extension, $D_{\text{bucket}}$. Excavation was deemed over when the bucket cylinder reached $D_{\text{bucket}}=420\ \rm mm$.

One notable aspect of these excavation experiments was that they were carried out by skilled operators under representative conditions in an operating quarry.  To this end, machine Operators A and B were instructed to obtain a full bucket of material and to avoid scraping the ground prior to entry; otherwise, they were allowed full reign over the excavation process, including inherent operator variability.

\begin{figure}
  \centering
  \includegraphics[width=0.85\linewidth]{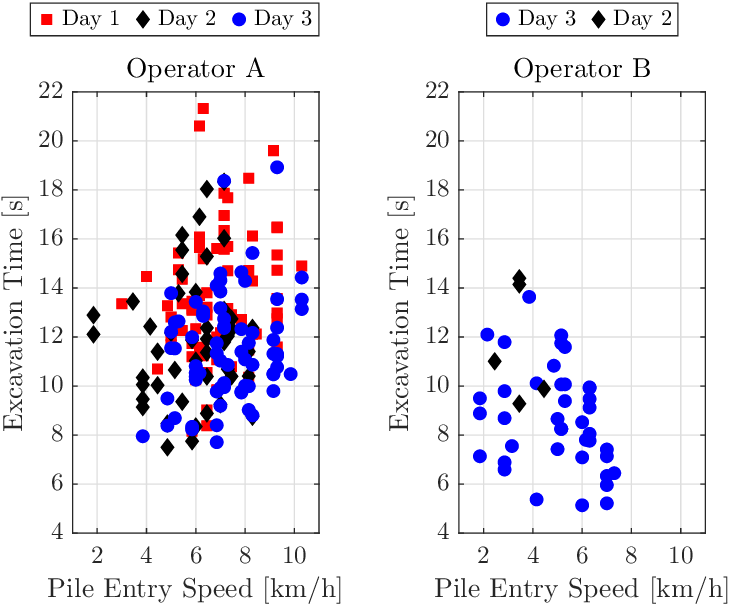}
  \caption{Evaluation of excavation time and vehicle speed at start.}
  \label{fig:dig-time-variability}
\end{figure}

Despite being given the same instructions, Operators A and B had different experience levels and behaved differently in their approach to filling the excavator's bucket at each pass.  Figure\ \ref{fig:dig-time-variability} shows the variability in excavation time (i.e., $\alpha_2-\alpha_1$) against entry speed into the rock pile for both operators.  Note the variability and that Operator B took a more conservative approach.  Furthermore, by inspecting Figure\ \ref{fig:dig-time-variability}, the majority of trials for Operator B were performed within 11~s, highlighting that signals past this point for Operator A likely occurred when the bucket had left the pile and therefore may not be informative for fragmentation estimation purposes.  Thus, $\alpha_2$ values were capped at 11~s. Determining when excavation effectively terminates—when additional excavation effort ceases to increase payload—is a practical challenge. The 11~s threshold established from Operator B's consistent performance provides a data-driven boundary between productive bucket-pile engagement and subsequent operator-specific material handling activities.

\begin{figure}
  \centering
  \includegraphics[width=0.85\linewidth]{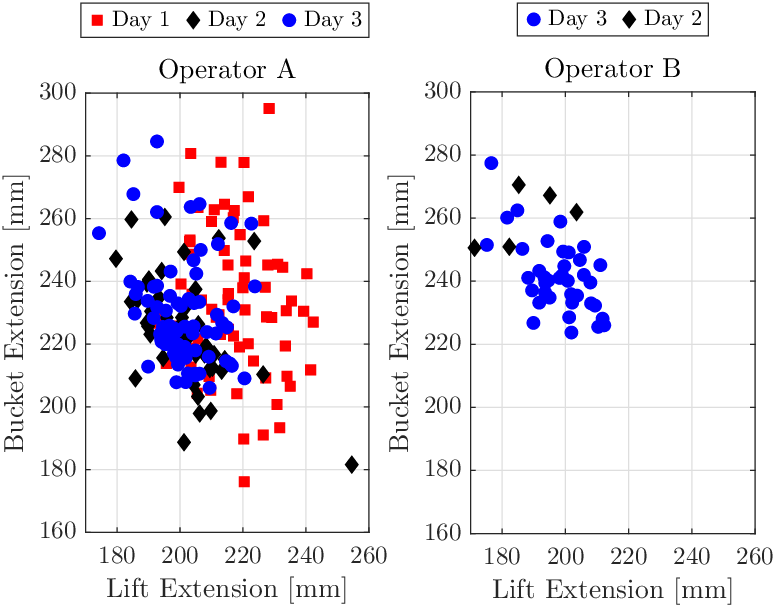}
  \caption{Variability of the lift and bucket cylinder extensions at entry.}
  \label{fig:cylinder-extension-variability}
\end{figure}

Figure\ \ref{fig:cylinder-extension-variability} illustrates operator variability by showing the distribution of the lift and bucket cylinder positions at entry.  After the first day of trials, Operator A became more comfortable with the task and more consistent. Operator B after their first day of trials, which started on Day 2 of experiments, increased the pile entry speed and quickened their excavation times.

\subsection{DATA PRE-PROCESSING}

\begin{figure}
  \centering
  \subfloat[Bucket cylinder.]{
    \includegraphics[width=0.99\columnwidth]{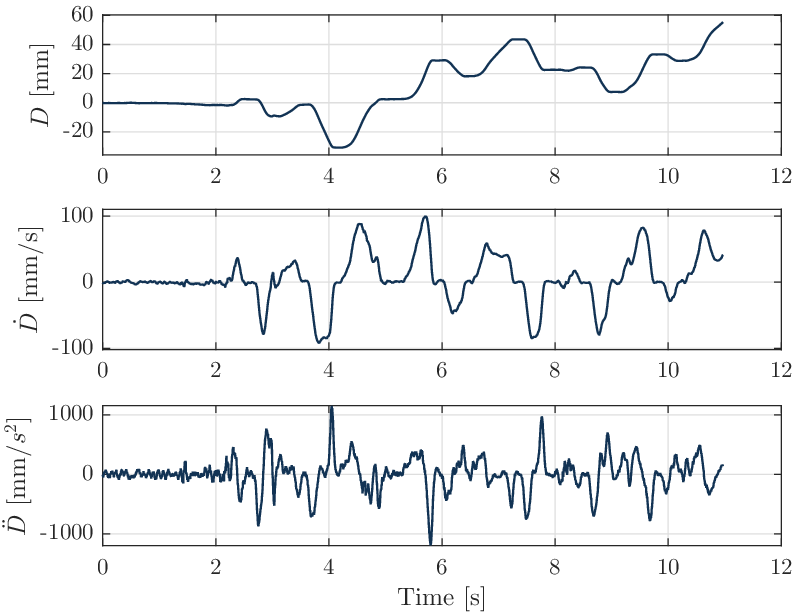}
  }
  \hfill
  \subfloat[Boom cylinder.]{
    \includegraphics[width=0.99\columnwidth]{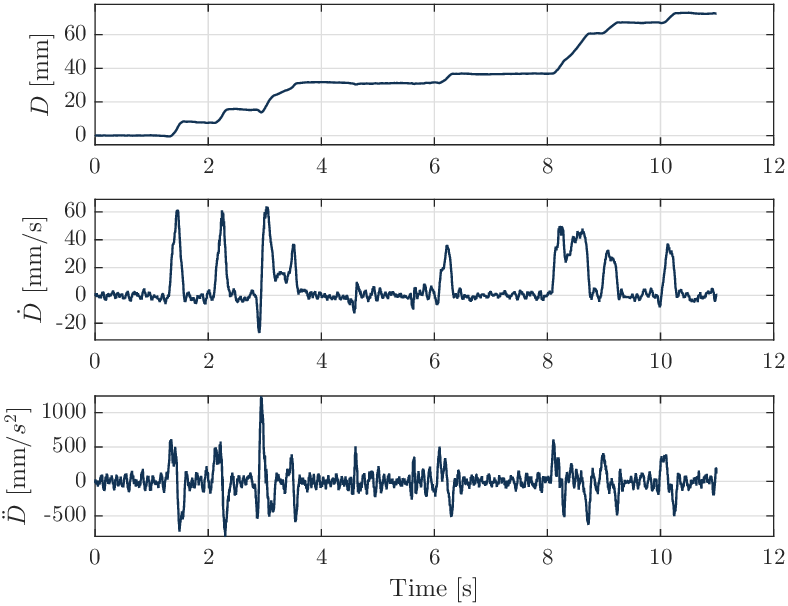}
  }
\caption{Example cylinder extension ($D$), velocity ($\dot{D}$), and acceleration ($\ddot{D}$) for Operator A excavating in pile 0/32. Cylinder extensions were zeroed at pile entry. The bucket cylinder exhibits higher frequency oscillations due to active probing during excavation, while the boom cylinder shows lower frequency actuation for gross positioning.}

  \label{fig:cylinder_info_0_32}
\end{figure}

\begin{figure}
  \centering
  \subfloat[Bucket cylinder.]{
    \includegraphics[width=0.99\columnwidth]{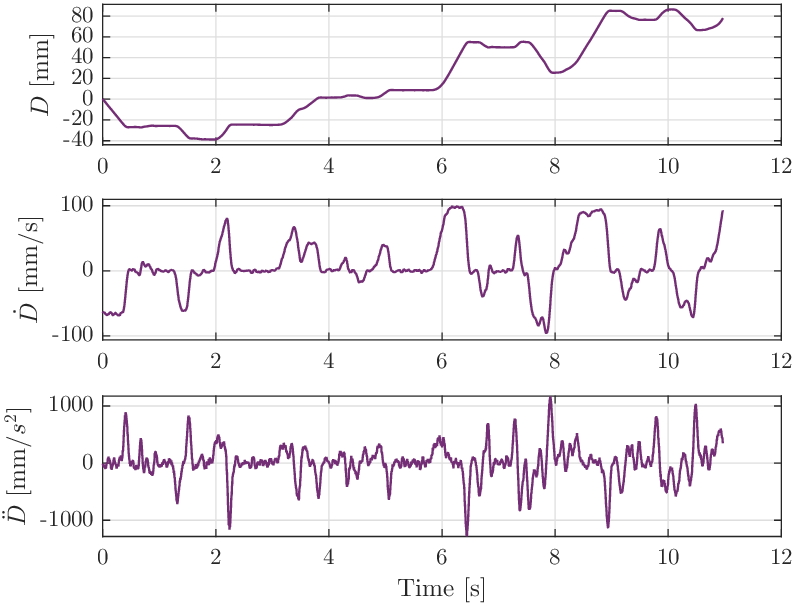}
  }
  \hfill
  \subfloat[Boom cylinder.]{
    \includegraphics[width=0.99\columnwidth]{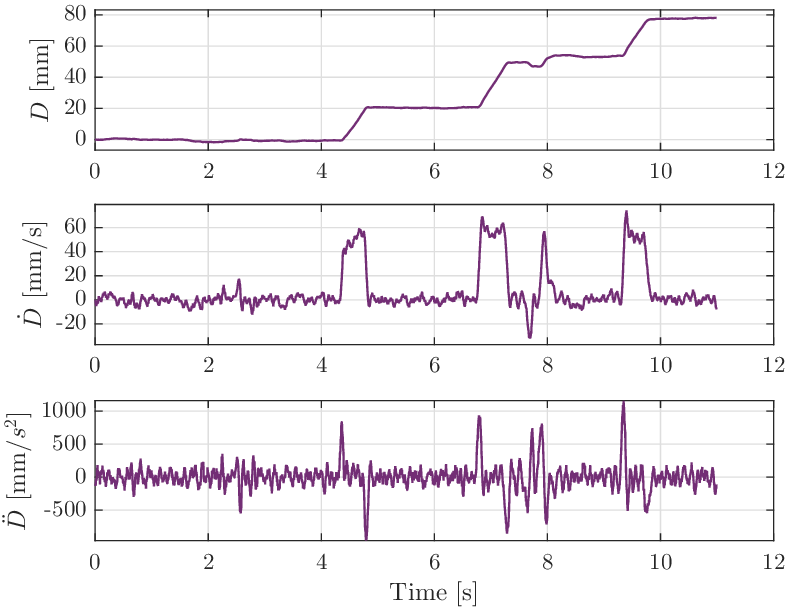}
  }
\caption{Example cylinder extension ($D$), velocity ($\dot{D}$), and acceleration ($\ddot{D}$) for Operator A excavating in pile 0/150. Cylinder extensions are zeroed at pile entry. The bucket cylinder exhibits higher frequency oscillations due to active probing during excavation, while the boom cylinder shows lower frequency actuation for gross positioning.}

  \label{fig:cylinder_info_0_150}
\end{figure}

Signals generated during excavations were processed using a high-pass filter.  A cut-off frequency $f_{k}=2$\ Hz for the boom IMU signals, and $f_{k}=4$\ Hz for the bucket IMU signals were decided, experimentally.  The selected bucket IMU cut-off was larger due to the oscillatory style of commanding the bucket cylinder from both operators. This difference in frequency can be observed through the bucket and boom cylinder positions, velocities, and accelerations, with examples provided in Figure\ \ref{fig:cylinder_info_0_32} and Figure\ \ref{fig:cylinder_info_0_150}.

This operator-injected low-frequency oscillation is not a result of bucket-rock interactions and, if not removed, would impact the wavelet analysis and the generated wavelet feature.

\subsection{MATERIAL CHARACTERIZATION}
\label{sec:results-material-characterization}

\begin{figure}
  \centering
  \includegraphics[width=\linewidth]{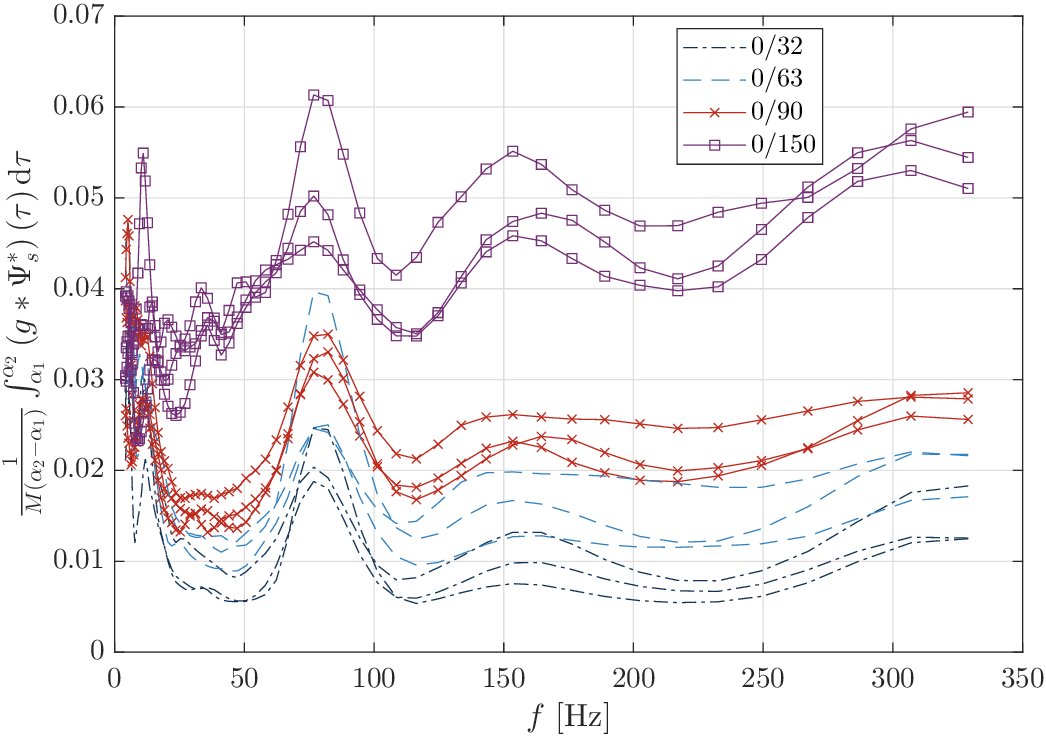}
  \caption{Example wavelet analysis output generated by using the bucket-mounted IMU 2 from Operator A's trials prior to integrating over the frequencies to obtain the wavelet feature $\zeta$.}
  \label{fig:fragx-features-no-integration}
\end{figure}

The primary contribution of this field report is to experimentally test the wavelet feature $\zeta$'s effectiveness at encapsulating changes in the rock pile's fragmentation characteristics.  Example wavelet analysis signals, as a function of frequency $f$, for trials conducted in different rock piles are shown in Figure\ \ref{fig:fragx-features-no-integration}.  The novelty of this methodology lies in its potential to provide an estimate of the relative mean particle size of different rock piles by using only proprioceptive data.  

\subsubsection{RELATIVE MEAN ROCK SIZE ESTIMATION RESULTS}
\label{sec:results-relative-mean-rock-size}

The $\zeta_i$ feature generated from interactions with a given rock pile $i$ can be compared against a reference pile's $\zeta_{\rm ref}$ value.  In this study, we tested both the 0/90 and 0/150 piles as reference piles.  It practice, as a potential product for industry practitioners, this could be done ahead of time as a calibration step.  The ratio $\zeta_i:\zeta_{\rm ref}$ was then used to automatically estimate the magnitude of pile $i$'s mean particle size $\bar{x}_i$ by using that of the reference pile mean particle size $\bar{x}_{\rm ref}$ as per \eqref{eq:wavelet-mean-ratios}. For comparison, the sieve and WipFrag\texttrademark\ mean particle size estimates were also normalized by the reference pile's mean particle size estimate (for each respective method) to allow for a critical evaluation of the proposed proprioceptive method against what is estimated by both a widely-used vision-based system and by sieve analysis.

\begin{figure}
  \centering
  \includegraphics[width=\linewidth]{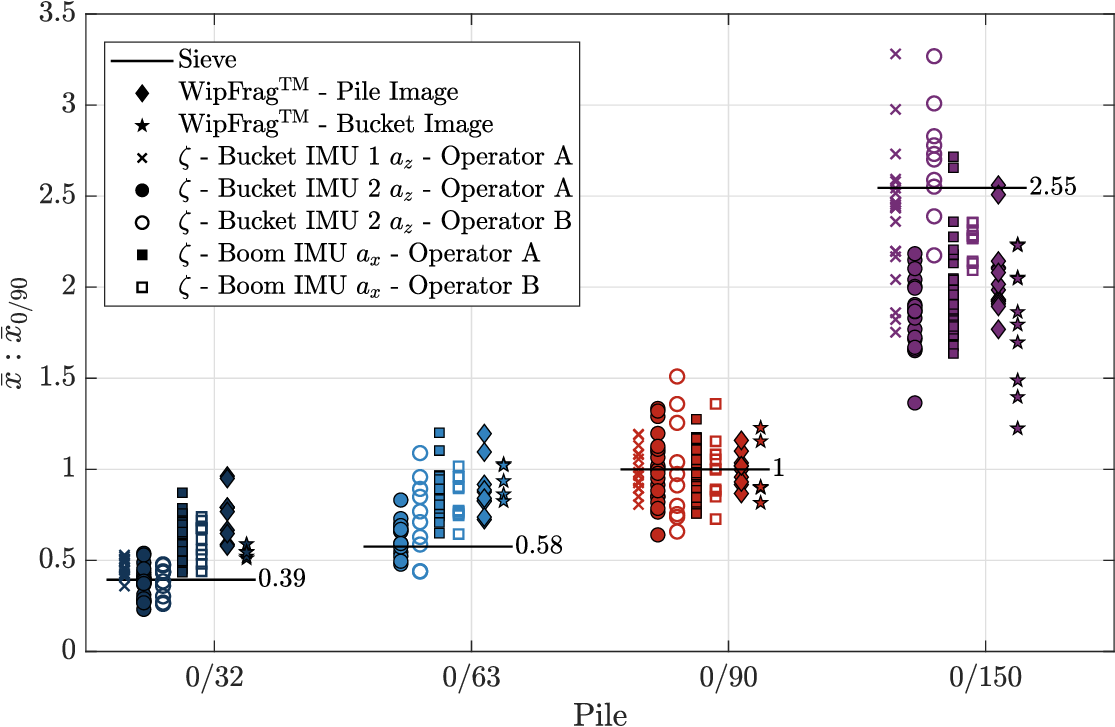}
  \caption{Relative mean particle size estimates from the signals generated from the boom and bucket IMUs and compared with sieve and WipFrag\texttrademark.  Pile 0/90 used as reference.}
  \label{fig:fragx-relative-size-results-no-1500-0-90-reference}
\end{figure}

\begin{figure}
  \centering
  \includegraphics[width=\linewidth]{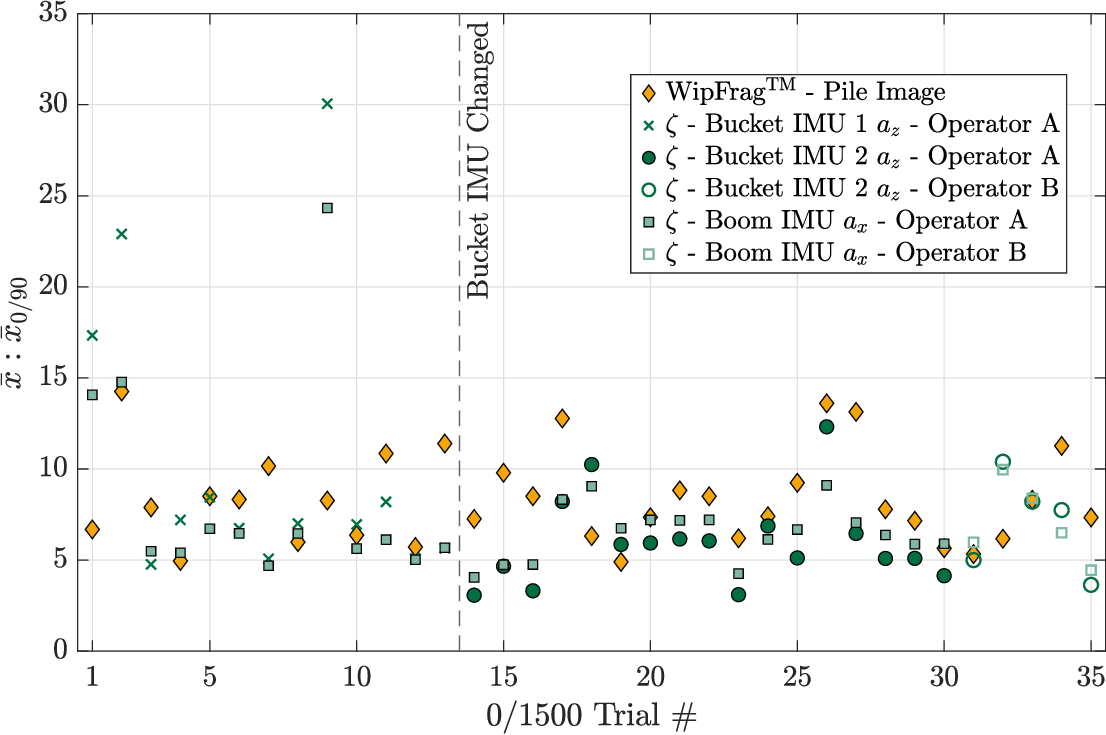}
  \caption{Relative mean particle size estimates for the 0/1500 pile for Operator A and compared with WipFrag\texttrademark.  Pile 0/90 used as reference.}
  \label{fig:fragx-relative-size-results-0-1500-0-90-reference}
\end{figure}

\begin{figure}
  \centering
  \includegraphics[width=\linewidth]{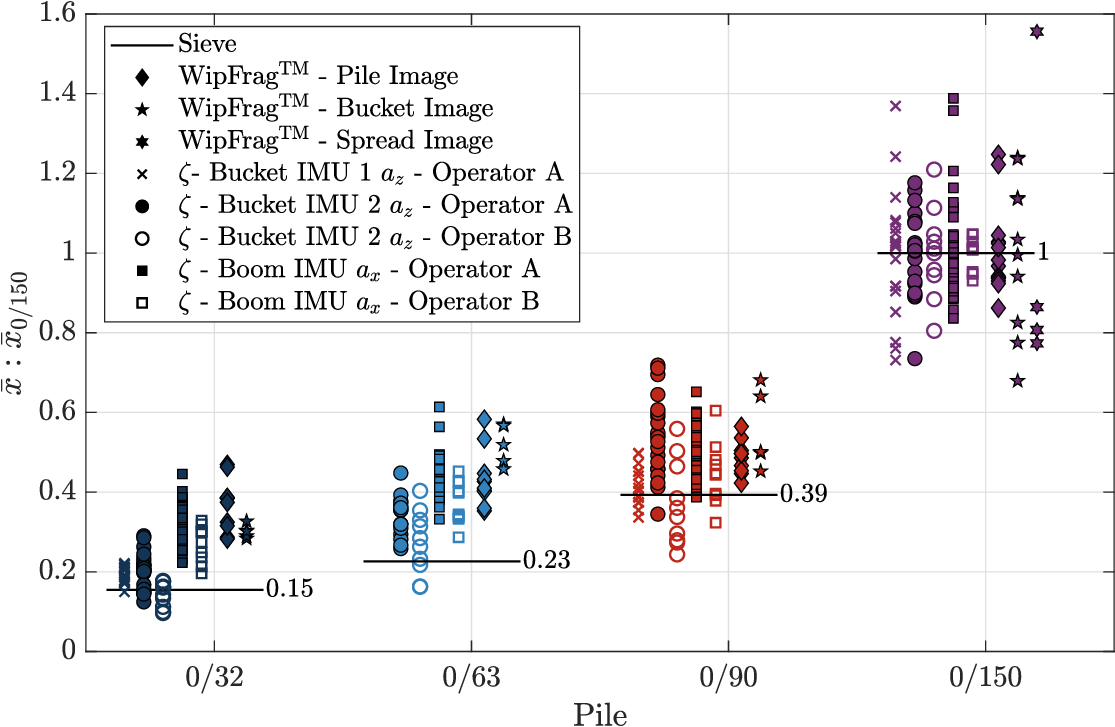}
  \caption{Relative mean particle size estimates from the signals generated from the boom and bucket IMUs and compared with sieve and WipFrag\texttrademark.  Pile 0/150 used as reference.}
  \label{fig:fragx-relative-size-results-no-1500-0-150-reference}
\end{figure}

\begin{figure}
  \centering
  \includegraphics[width=\linewidth]{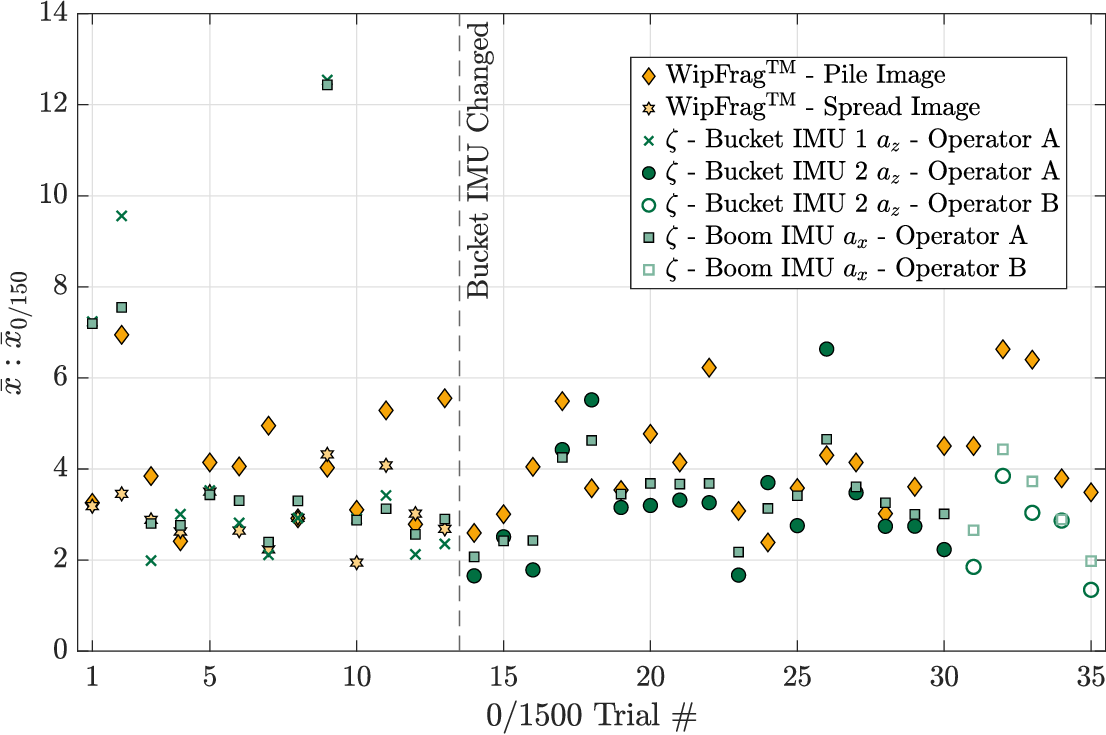}
  \caption{Relative mean particle size estimates for the 0/1500 pile for Operator A and compared with WipFrag\texttrademark.  Pile 0/150 used as reference.}
  \label{fig:fragx-relative-size-results-0-1500-0-150-reference}
\end{figure}

Figures\ \ref{fig:fragx-relative-size-results-no-1500-0-90-reference}, \ref{fig:fragx-relative-size-results-0-1500-0-90-reference} and Figures\ \ref{fig:fragx-relative-size-results-no-1500-0-150-reference}, \ref{fig:fragx-relative-size-results-0-1500-0-150-reference} allow for a comparative evaluation of estimating the relative mean particle size estimates by sieve analysis, WipFrag\texttrademark and our method and an assessment of the variability of the estimates of each technique.  We tested both the 0/90 and 0/150 piles\footnote{Note that WipFrag\texttrademark\ estimates for the ``spread'' images (see, e.g., Figure\ \ref{fig:0-150-spread-example}) are included with pile 0/150 for completeness.  No ``spread'' images were captured for the 0/90 pile.}, respectively, as the chosen ``calibration'' reference.  Means and standard deviations for the respective data collected during full-scale excavations are also provided in \cref{tbl:material-characterization-summary}. As can be seen from the data, the proposed proprioceptive-only approach distinguishes the relative mean particle size with similar accuracy to that of sieving.  However, some difficulty arises (i.e., overlap) in distinguishing pile 0/63, which is true across all of the tested methods and is clearly due to its very similar size distribution to that of 0/90, as can be observed in Figure\ \ref{fig:semi-homogeneous-size-estimates}.  In other words, there are limits to the resolution at which this technique (and similarly, visual and sampling) can estimate fragmentation parameters.

\begin{table*}
    \caption{Summary of material characterization results $\bar{x}:\bar{x}_{\rm ref}$ for the different methods studied by full-scale experiments shown in Figure\ \ref{fig:fragx-relative-size-results-no-1500-0-90-reference}, Figure\ \ref{fig:fragx-relative-size-results-0-1500-0-90-reference} and Figure\ \ref{fig:fragx-relative-size-results-no-1500-0-150-reference}, Figure\ \ref{fig:fragx-relative-size-results-0-1500-0-150-reference}.  The notation $\mu \pm \sigma$ denotes the mean and one standard deviation of the data.}
    \label{tbl:material-characterization-summary}
    \centering
    \begin{tabular}{l|ccccc}
        \toprule
        \bf Method & \bf 0/32 & \bf 0/63 & \bf 0/90 & \bf 0/150  & \bf 0/1500 \\ \midrule
        \bf Figure\ \ref{fig:fragx-relative-size-results-no-1500-0-90-reference} & $\bar{x}_{0/32}:\bar{x}_{0/90}$ & $\bar{x}_{0/63}:\bar{x}_{0/90}$ & $\bar{x}_{0/90}:\bar{x}_{0/90}$ & $\bar{x}_{0/150}:\bar{x}_{0/90}$ &  \\ \midrule
        Sieve & 0.39 & 0.58 & 1 (ref) & 2.55 & - \\
        WipFrag\texttrademark\ - Pile Image & 0.75$\pm$0.13 & 0.89$\pm$0.15 & 1$\pm$0.09 (ref)  & 2.05$\pm$0.22 & - \\
        WipFrag\texttrademark\ - Bucket Image & 0.54$\pm$0.03 & 0.94$\pm$0.09 & 1$\pm$0.18 (ref) & 1.80$\pm$0.35 & - \\
        Bucket IMU 1 (Operator A) & 0.46$\pm$0.05 & - & 1$\pm$0.12 (ref) & 2.40$\pm$0.36 & - \\
        Bucket IMU 2 (Operator A) & 0.38$\pm$0.08 & 0.63$\pm$0.10 & 1$\pm$0.19 (ref) & 1.86$\pm$0.20 & - \\
        Bucket IMU 2 (Operator B) & 0.38$\pm$0.08 & 0.74$\pm$0.22 & 1$\pm$0.29 (ref) & 2.70$\pm$0.31 & - \\
        Boom IMU (Operator A) & 0.62$\pm$0.10 & 0.86$\pm$0.11 & 1$\pm$0.13 (ref) & 1.96$\pm$0.24 & - \\
        Boom IMU (Operator B) & 0.61$\pm$0.10 & 0.84$\pm$0.12  & 1$\pm$0.18 (ref) & 2.24$\pm$0.09 & - \\ \midrule
        \bf Figure\ \ref{fig:fragx-relative-size-results-0-1500-0-90-reference} & & & $\bar{x}_{0/90}:\bar{x}_{0/90}$ & & $\bar{x}_{0/1500}:\bar{x}_{0/90}$ \\ \midrule
        WipFrag\texttrademark\ - Pile Image & - & - & 1$\pm$0.09 (ref) & - & 8.35$\pm$2.51 \\
        Bucket IMU 1 (Operator A) & - & - & 1$\pm$0.12 (ref) & - &  10.42$\pm$7.94 \\
        Bucket IMU 2 (Operator A) & - & - & 1$\pm$0.19 (ref) & - &  5.98$\pm$2.45 \\
        Bucket IMU 2 (Operator B) & - & - & 1$\pm$0.29 (ref) & - &  7.00$\pm$2.69  \\
        Boom IMU (Operator A) & - & - & 1$\pm$0.13 (ref) & - &  7.38$\pm$4.01 \\
        Boom IMU (Operator B) & - & - & 1$\pm$0.18 (ref) & - &  7.05$\pm$2.15 \\ \midrule
        \bf Figure\ \ref{fig:fragx-relative-size-results-no-1500-0-150-reference} & $\bar{x}_{0/32}:\bar{x}_{0/150}$ & $\bar{x}_{0/63}:\bar{x}_{0/150}$ & {$\bar{x}_{0/90}:\bar{x}_{0/150}$} & {$\bar{x}_{0/150}:\bar{x}_{0/150}$} & \\ \midrule
        Sieve & 0.15 & 0.23 & 0.39 & 1 (ref) & - \\
        WipFrag\texttrademark\ - Pile Image & 0.37$\pm$0.07 & 0.44$\pm$0.07 & 0.49$\pm$0.04 &  1$\pm$0.11 (ref) & - \\
        WipFrag\texttrademark\ - Bucket Image & 0.30$\pm$0.02 & 0.52$\pm$0.05 & 0.55$\pm$0.10 &  1$\pm$0.19 (ref) & - \\
        Bucket IMU 1 (Operator A) & 0.19$\pm$0.02 & - & 0.42$\pm$0.05 & 1$\pm$0.15 (ref)  & - \\
        Bucket IMU 2 (Operator A) & 0.20$\pm$0.04 & 0.34$\pm$0.05 & 0.54$\pm$0.10 &  1$\pm$0.11 (ref) & - \\
        Bucket IMU 2 (Operator B) & 0.14$\pm$0.03 & 0.27$\pm$0.08 & 0.37$\pm$0.11 & 1$\pm$0.11 (ref) & - \\
        Boom IMU (Operator A) & 0.32$\pm$0.05 & 0.44$\pm$0.06 & 0.51$\pm$0.07 & 1$\pm$0.12 (ref) & - \\
        Boom IMU (Operator B) & 0.27$\pm$0.05 & 0.37$\pm$0.05 & 0.44$\pm$0.08 & 1$\pm$0.04 (ref) & - \\ \midrule
        \bf Figure\ \ref{fig:fragx-relative-size-results-0-1500-0-150-reference} & & & & $\bar{x}_{0/150}:\bar{x}_{0/150}$ & $\bar{x}_{0/1500}:\bar{x}_{0/150}$ \\ \midrule
        WipFrag\texttrademark\ - Pile Image & - & - & - & 1$\pm$0.11 (ref) & 4.07$\pm$1.22 \\
        WipFrag\texttrademark\ - Spread Image & - & - & - & 1$\pm$0.37 (ref) & 3.04$\pm$0.68 \\
        Bucket IMU 1 (Operator A) & - & - & - & 1$\pm$0.15 (ref) & 4.35$\pm$3.32   \\
        Bucket IMU 2 (Operator A) & - & - & - & 1$\pm$0.11 (ref) & 3.22$\pm$1.32  \\
        Bucket IMU 2 (Operator B) & - & - & - & 1$\pm$0.11 (ref) & 2.59$\pm$0.99  \\
        Boom IMU (Operator A) & - & - & - & 1$\pm$0.12 (ref) & 3.77$\pm$2.05  \\
        Boom IMU (Operator B) & - & - & - & 1$\pm$0.04 (ref) & 3.14$\pm$0.96  \\
        \bottomrule
    \end{tabular}
\end{table*}

\subsubsection{APPLICATION TO CLASSIFICATION}
\label{sec:results-smaller-or-bigger}

In this section, we briefly illustrate one possible application of the proposed proprioceptive fragmentation estimation method---namely, that of pile classification, which is of practical significance for mining operations. Beyond immediate operational use, this classification capability could also inform excavation strategies for autonomous systems, where post-excavation material classification provides feedback for tuning algorithm parameters. However, experimental validation of this potential application remains as future work for autonomous excavation systems.

Consider the objective of classifying excavated buckets of fragmented rock as either \textit{smaller} or \textit{larger} in comparison to a reference rock pile.  This is useful in mining, where materials that are too large may need additional breaking.  Suppose that the reference pile's estimates are assumed to be normally distributed; i.e., $\zeta_{\rm ref}\sim\mathcal{N}(\mu_{\rm ref},\sigma^{2}_{\rm ref})$.  Although there are a number of ways to perform classification, one simple approach could be to consider a given pile $i$ as \textit{smaller} than the reference pile if, for example,
\begin{equation}
    z_i = \frac{\zeta_i-\mu_{\rm ref}}{\sigma_{\rm ref}} < -z_{p}
\end{equation}
and \emph{larger} if $z_i > z_p$, where $p\in\{0.90, 0.95, 0.99\}$ are the probability bounds (e.g., $z_{0.90}=1.645$, etc.). \cref{tbl:FragX-classification-smaller-larger} provides the classification accuracy for each signal source and for the two reference piles. With pile 0/90 as reference, the 0/63 pile was difficult to distinguish, which could explain the lower classification accuracy observed in \cref{tbl:FragX-classification-smaller-larger}. Nevertheless, these results suggest that the wavelet feature achieves classification performance comparable to vision-based fragmentation analysis while enabling bucket-by-bucket assessment immediately after each excavation cycle in dark and dusty conditions.

\begin{table*}
    \caption{Classification of rock piles as either \textit{smaller} or \textit{larger} than a reference rock pile.  Reference pile was either 0/90 or 0/150.}
    \label{tbl:FragX-classification-smaller-larger}
    \centering
    \begin{tabular}{l|ccc|cc|c}
    \toprule
    \bf Reference & \multicolumn{3}{c|}{\bf Operator A} & \multicolumn{2}{c|}{\bf Operator B} & \multicolumn{1}{c}{\bf WipFrag\texttrademark} \\
    \bf Pile, $z_p$ & \multicolumn{1}{c}{\bf Bucket IMU 1} & \multicolumn{1}{c}{\bf Bucket IMU 2} & \bf Boom IMU & \multicolumn{1}{c}{\bf Bucket IMU 2} & \bf Boom IMU & \bf Pile Images \\
     \midrule
    0/90, $z_{0.90}$ & \multicolumn{1}{c}{96 \%} & \multicolumn{1}{c}{100 \%} & 87 \% & \multicolumn{1}{c}{82 \%} & 76 \% & 89 \% \\
    0/90, $z_{0.95}$ & \multicolumn{1}{c}{91 \%} & \multicolumn{1}{c}{100 \%} & 83 \% & \multicolumn{1}{c}{69 \%} & 69 \% & 85 \% \\
    0/90, $z_{0.99}$ & \multicolumn{1}{c}{84 \%} & \multicolumn{1}{c}{100 \%} & 76 \% & \multicolumn{1}{c}{56 \%} & 62 \% & 76 \% \\
    0/150, $z_{0.90}$ & \multicolumn{1}{c}{100 \%} & \multicolumn{1}{c}{100 \%} & 100 \% & \multicolumn{1}{c}{100 \%} & 100 \%  & 100 \% \\
    0/150, $z_{0.95}$ & \multicolumn{1}{c}{100 \%} & \multicolumn{1}{c}{100 \%} & 100 \% & \multicolumn{1}{c}{100 \%} & 100 \%  & 100 \% \\
    0/150, $z_{0.99}$ & \multicolumn{1}{c}{100 \%} & \multicolumn{1}{c}{100 \%} & 100 \% & \multicolumn{1}{c}{100 \%} & 100 \%  & 100 \% \\
    \bottomrule
    \end{tabular}%
\end{table*}

\section{ADDITIONAL DISCUSSION}
\label{sec:discussion}

In this section we discuss some further observations from our field experiments about the studied proprioceptive technique for rock pile characterization.

\subsection{Ground Truth Challenges}

Although one might consider sieving as the obvious ground truth to which we should compare the proposed fragmentation estimation technique, the truth is that only a ``representative'' sample is used for sieve analysis.  Furthermore, this sample significantly influences the model estimate.  For example, consider the 0/150 pile used in our field experiments.  The sieve estimates $P(x)$ at $x=90$ mm, $x=125$ mm, and $x=180$ mm are below the lower bounds given in standard TDOK 2013:0530v3, to which the pile 0/150 should conform \cite{SS-EN-933}.  However, if instead the regression used $P(90)=80 \%$, $P(125)=90 \%$ and $P(180)$ was excluded, the model's mean particle size $\bar{x}_{0/150}$ changes from 84 to 72 mm and $n$ from 0.8519 to 0.9160.  This highlights one of the challenges in evaluating fragmentation estimates, where consistent and accurate ground truth estimates of $\bar{x}$ are difficult to obtain.

\subsection{Lift Cylinder Limitations}

\begin{figure}
  \centering
  \includegraphics[width=\linewidth]{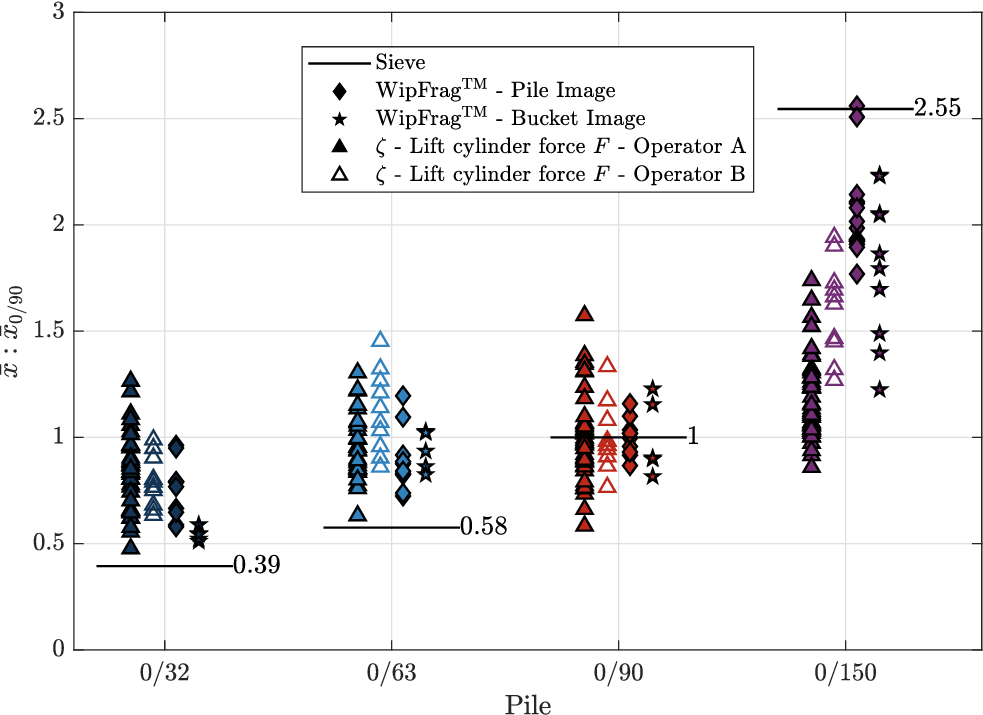}
  \caption{Relative mean particle size estimates using the lift cylinder forces compared with sieve and WipFrag\texttrademark.  Pile 0/90 used as reference.}
  \label{fig:fragx-relative-size-results-lift-forces}
\end{figure}

In the development of the proposed technique \cite{ArtFerMar-AIM,ArtanU-PhD2022}, lift cylinder forces were hypothesized as a potential signal for fragmentation analysis. However, in that early exploratory work, the lift cylinders served only as a passive sensor because they were never actuated and their pressures were sampled at only 20 Hz. In the field experiments presented by this paper, the sampling rate was increased to 250 Hz and expert operators were allowed to (more realistically) use the lift cylinders during digging trials. However, in this case, Figure\ \ref{fig:fragx-relative-size-results-lift-forces} reveals that lift cylinder forces are not practically useful for differentiating rock piles by wavelet analysis.

This limitation is not due to sensor accuracy---the pressure transducers ($\pm$0.25\% full scale) and position sensors ($\pm$0.05\% linearity) provide sufficient measurement precision. We hypothesize the limitation stems from operators actively commanding the lift cylinders for bucket positioning during excavation, as shown in Figure\ \ref{fig:cylinder_info_0_32} and Figure\ \ref{fig:cylinder_info_0_150}. This active actuation likely introduces hydraulic damping that attenuates the high-frequency oscillations that the proprioceptive wavelet-based approach relies on to characterize bucket-rock interactions. This hypothesis is supported by prior work \cite{ArtanU-PhD2022,egli2024reinforcement} on autonomous excavators, where lift cylinders served effectively as passive force sensors when not actively commanded during material loading. Future work should investigate whether lift cylinders can provide useful fragmentation information when operated as passive sensors, potentially by analyzing excavation phases where they are not actuated or by employing higher sampling rates to capture dynamics before damping effects dominate.

\subsection{Operator Variability and Autonomous Applications}

\begin{figure}
  \centering
  \includegraphics[width=\linewidth]{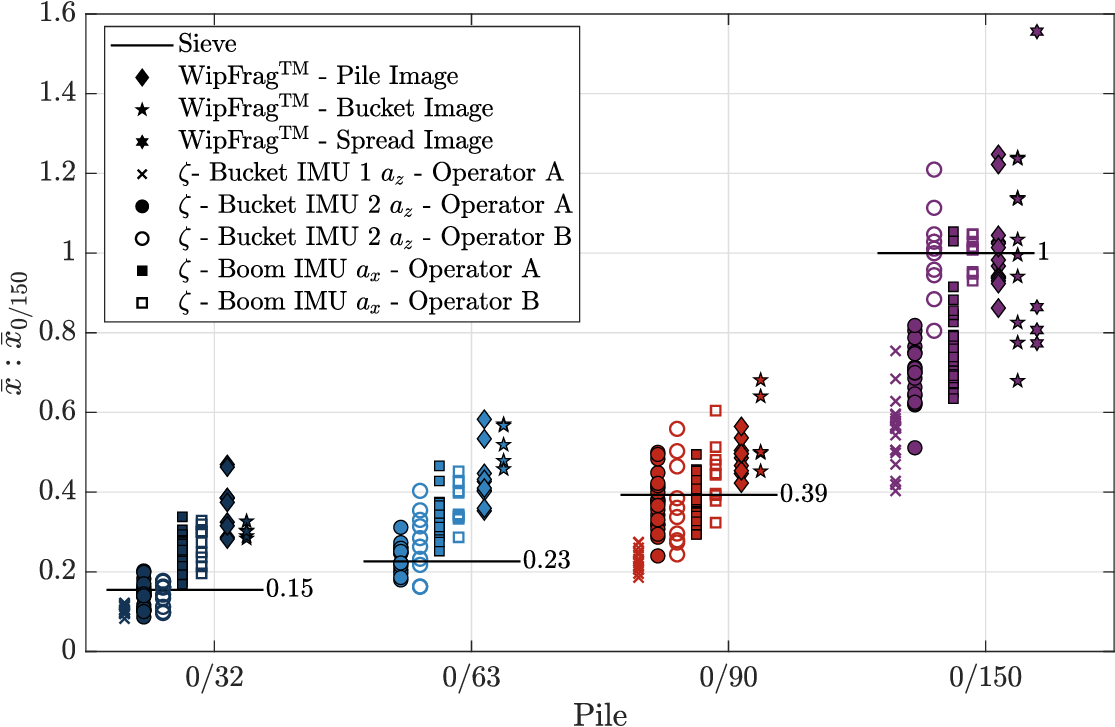}
  \caption{Relative mean particle size estimates and the mean $\zeta$ generate from Operator B's excavation trials in the 0/150 pile as reference.}
  \label{fig:fragx-relative-size-results-no-1500-0-150-isak-reference}
\end{figure}

One important question that arises is whether the proposed technique is highly dependent or unduly influenced by the behavior of the operator. Figure\ \ref{fig:fragx-relative-size-results-no-1500-0-150-isak-reference} focuses on Operator B and shows what happens if only that operator's excavation trials are used as the reference (e.g., as a calibration by a single operator). It is worth reiterating that Operator A's approach was significantly more aggressive in comparison with that of Operator B.  As such, the results presented by this figure do suggest that individual operators may have some influence on the achievable accuracy.  This is significant because it suggests that the proposed proprioceptive approach to rock pile fragmentation estimation is ideally suited to autonomous excavators \cite{Marshall2008,Dobson2017,AtlasCopcoAutoLoad} due to the consistency of behaviors that results from the use of automated machines.

\subsection{Future Work and Study Limitations}

This field campaign focused on validating relative average particle size estimation using full-scale equipment under operational conditions---a critical prerequisite for deployment and use with robotic excavation systems. The wavelet features may also contain information about particle size distributions, beyond merely the relative average size metric. Extension of the analysis to extract these more detailed distribution characteristics is ongoing.

The experiments were conducted using blasted rock and crushed aggregate from a single quarry. Laboratory analysis of samples indicated a material density of $\rho = 2.63$~t$/$m$^{3}$, and uniform density was assumed across all piles, though actual spatial variations may exist and could contribute to the observed variability in results. In practice, mining operations encounter additional variability in material properties (density, moisture content, friction, cohesion, packing fraction) across different geological formations that affect excavation and thus particle size estimation. This limitation affects all methods: vision-based systems similarly assume uniform density, while sieve analysis can measure density from samples but not spatial distribution. Future work should investigate robustness to material property variations and the need for site-specific calibration strategies.

\section{CONCLUSION}
\label{sec:conclusion}

This field report presents the results from full-scale field experiments to evaluate a novel methodology for estimating the mean rock size of excavation materials by using only proprioceptive sensing (i.e., force and IMU-measured acceleration data, in this case) acquired from an excavator working in multiple rock piles of varying characteristics. The method depends only on proprioceptive sensing and thus works in dark, dusty and harsh environments that often limit the effective use of exteroceptive sensor data.  Our field experiments were conducted at an operating quarry in Sweden, where reliable estimates of the material characteristics could be obtained through sampling and sieving, as well as by comparison with a vision-based commercial product to assess the accuracy of the proposed technology.  

Our field experiments conclude that the proposed wavelet-based analysis technique is able to distinguish rock pile types by (relative) mean particle size on par with the accuracy and uncertainty associated with an industry-standard vision-based system (i.e., WipFrag\texttrademark), as well as by manually sampling and sieving rock particles.  The results also show that the behavior of manual operators may have some effect on the results---particularly if the system is calibrated using an excavation style that is significantly different from that used during use.  As such, the proposed technology is ideally suited for and supports the use by robotic excavation systems that provide consistent algorithmic approaches to autonomous excavation.

\section*{ACKNOWLEDGMENT}

Special thanks to Leif Gustafsson and Isak Östås at Epiroc Rock Drills AB for extensive field support. Many thanks to Mattias Pettersson, Katarina Öquist, Jan Gustafsson, Petter Forss and Christer Boman for fruitful discussions.  Many thanks to NCC AB for access to the active Eker quarry, where the field work presented in this paper was undertaken.  The authors are also extremely grateful to Thomas Palangio at WipWare, Inc.\ for providing us with access to their industry standard WipFrag\texttrademark\ software for research purposes, and for truly useful discussions about the technical challenges associated with different methods for estimating particle size distributions.


\bibliography{References}
\bibliographystyle{IEEEtran}


\begin{IEEEbiography}
[{\includegraphics[width=1in,height=1.25in,clip,keepaspectratio]{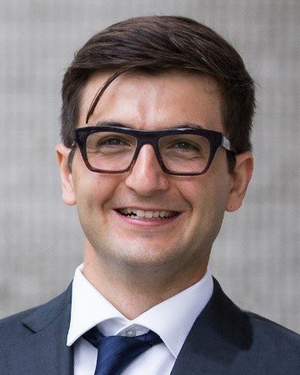}}]
{UNAL ARTAN}~received the Ph.D.\ degree in mining engineering from Queen's University, Kingston, ON Canada in 2022, the M.A.Sc.\ degree in aerospace engineering from Carleton University, Ottawa, ON Canada in 2010, and the B.Sc.Eng.\ degree in engineering physics from Queen's University, Kingston, ON Canada in 2007.

He is currently a Researcher with the Center for Applied Autonomous Sensor Systems (AASS) at \"{O}rebro University, \"{O}rebro, Sweden. His research focuses on learning from physical interactions to generate actionable insights for heavy machinery in the mining and construction industries. His prior industry work includes developing navigation and mapping solutions for space robotics and mining automation.

Dr.\ Artan was a recipient of the Queen Elizabeth II Graduate Scholarship in Science and Technology and the SAG Conference Award Foundation Student Award.

\end{IEEEbiography}





\begin{IEEEbiography}
[{\includegraphics[width=1in,height=1.25in,clip,keepaspectratio]{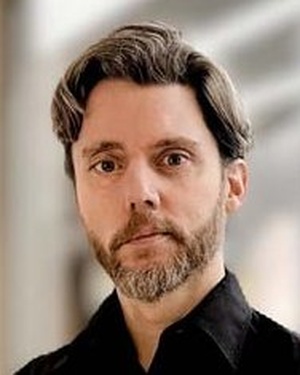}}]
{MARTIN MAGNUSSON}~received the Ph.D.\ degree in Computer Science from Örebro University in 2009 and the M.Sc.\ degree in Computer Science from Uppsala University in 2004. 

He is a Professor of Computer Science at Örebro University, where he currently leads the AASS research environment and its Robot Navigation and Perception Lab. His research focuses on robust perception, localisation, and 3D mapping for autonomous robots, with particular emphasis on reliability, uncertainty modelling, and introspective perception. His work spans domains such as autonomous vehicles and field \& service robotics, and he has led multiple nationally and internationally funded research projects in collaboration with academia, industry, and public stakeholders.

Prof.\ Magnusson is Associate Editor of the International Journal of Robotics Research (IJRR) and former Associate Editor of IEEE Robotics and Automation Letters (RA-L), the International Conference on Robotics and Automation (ICRA), and the International Conference on Intelligent Robots and Systems (IROS).
\end{IEEEbiography}

\begin{IEEEbiography}
[{\includegraphics[width=1in,height=1.25in,clip,keepaspectratio]{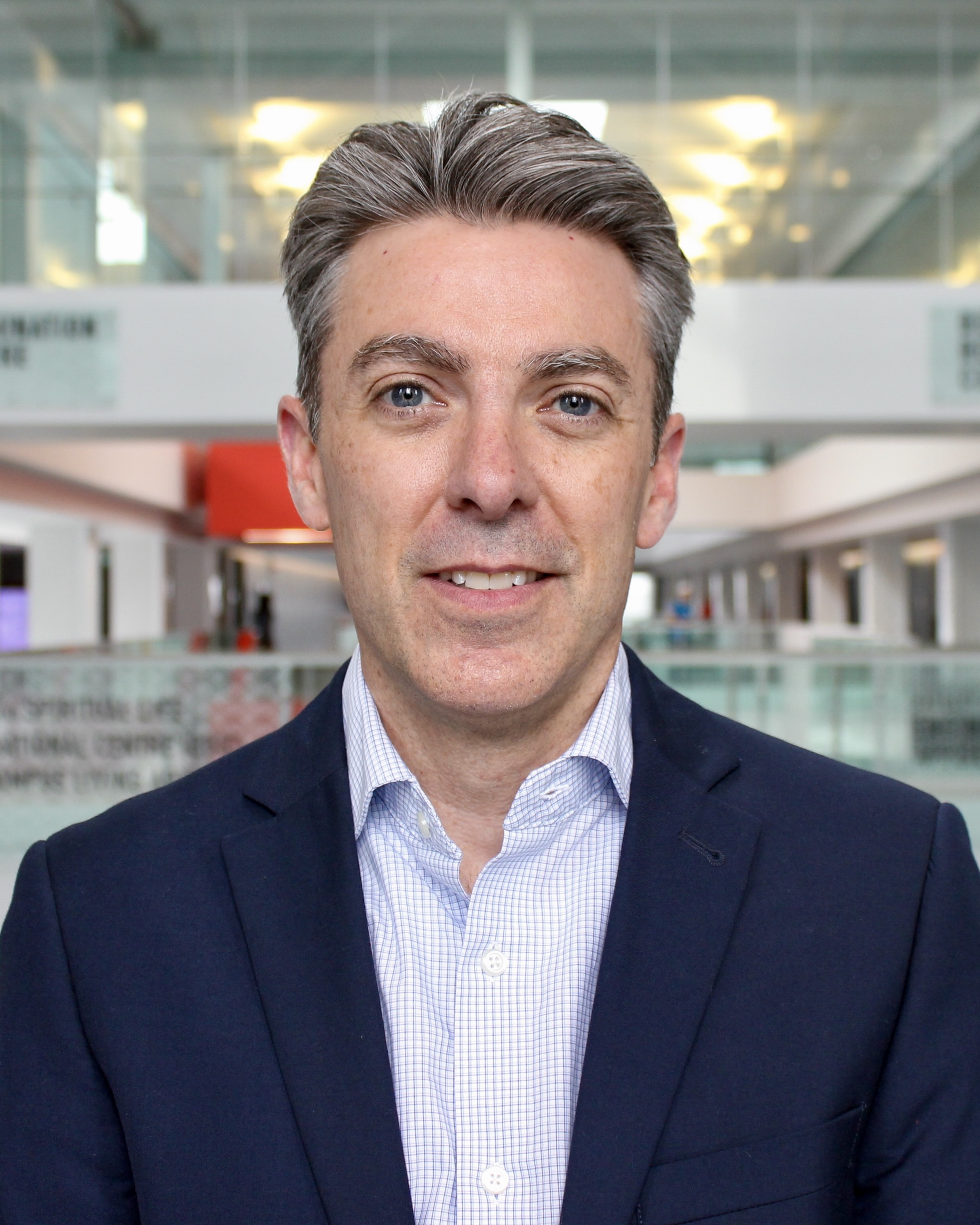}}]
{JOSHUA A.~MARSHALL}~(Senior Member, IEEE), was born in Iqaluit, NU Canada. He received the Ph.D.\ degree in electrical and computer engineering from the University of Toronto, Toronto, ON Canada in 2005, the M.Sc.\ degree in mechanical engineering in 2001, and the B.Sc.(Hons.)\ degree in mine-mechanical engineering in 1999, both from Queen's University, Kingston, ON Canada.

He is currently a Professor of mechatronics and robotics engineering in the Stephen J.R.\ Smith Faculty of Engineering and Applied Science at Queen's University, Kingston, ON Canada, where he directs the Offroad Robotics research group and is a faculty member at the Ingenuity Labs Research Institute, at which he previously served as founding Director. In 2016-17, he held the position of KKS Visiting Professor at the Center for Applied Autonomous Sensor Systems (AASS) at \"{O}rebro University, Sweden.  Prior to joining Queen's he was an Assistant Professor of mechanical and aerospace engineering at Carleton University, and served as control systems engineer at the Canadian robotics firm Macdonald, Dettwiler and Associates (MDA).  His research primarily focuses on autonomous mobile robot applications in mining, space, marine, and defence, as well as in other harsh-environments.

Prof.\ Marshall serves as an Associate Editor for the International Journal of Robotics Research (IJRR), with a concentration on dataset papers, and is the Cluster Chair for Field Robotics at the IEEE Robotics and Automation Society. He is a licensed Professional Engineer in the Province of Ontario, Canada and was awarded the 2025 Engineering Medal for R\&D by the Ontario Society of Professional Engineers (OSPE).

\end{IEEEbiography}

\vfill\pagebreak

\end{document}